\documentclass[acmtog, authorversion]{acmart}

\usepackage{booktabs} % For formal tables
\usepackage[ruled]{algorithm2e} % For algorithms

\SetAlFnt{\small}
\SetAlCapFnt{\small}
\SetAlCapNameFnt{\small}
\SetAlCapHSkip{0pt}
\IncMargin{-\parindent}

\setcopyright{none}

\usepackage{caption}
\usepackage{subcaption}
\usepackage{fix2col}
\usepackage{xcolor}
\newcommand{\reffig}[1]{Fig.~\ref{fig:#1}}
\newcommand{\refsec}[1]{Sec.~\ref{sec:#1}}

\newcommand{\refeqshort}[1]{(\ref{eq:#1})}

\newcommand{\lblfig}[1]{\label{fig:#1}}
\newcommand{\lblsec}[1]{\label{sec:#1}}
\newcommand{\lbleq}[1]{\label{eq:#1}}

\begin{document}
% Title portion
\title{Light Field Video Capture Using a Learning-Based Hybrid Imaging System} 
\author{Ting-Chun Wang}
\orcid{0000-0002-1522-2381}
\affiliation{%
  \institution{University of California, Berkeley}
  \city{Berkeley}
  \state{CA}
  \postcode{94720}
  \country{USA}
} 
\author{Jun-Yan Zhu}
\affiliation{%
  \institution{University of California, Berkeley}
  \department{Electrical Engineering and Computer Science}
  \city{Berkeley}
  \state{CA}
  \postcode{94720}
}
\author{Nima Khademi Kalantari}
\affiliation{%
  \institution{University of California, San Diego}
  \city{San Diego}
  \state{CA}
  \postcode{92093}
}
\author{Alexei A.\ Efros}
\affiliation{%
  \institution{University of California, Berkeley}
  \department{Electrical Engineering and Computer Science}
  \city{Berkeley}
  \state{CA}
  \postcode{94720}
}
\author{Ravi Ramamoorthi}
\affiliation{%
  \institution{University of California, San Diego}
  \city{San Diego}
  \state{CA}
  \postcode{92093}
}

\setlength{\abovedisplayskip}{3.5pt}
\setlength{\belowdisplayskip}{3.5pt}

\begin{teaserfigure}
   \centering
   \includegraphics[width=\linewidth]{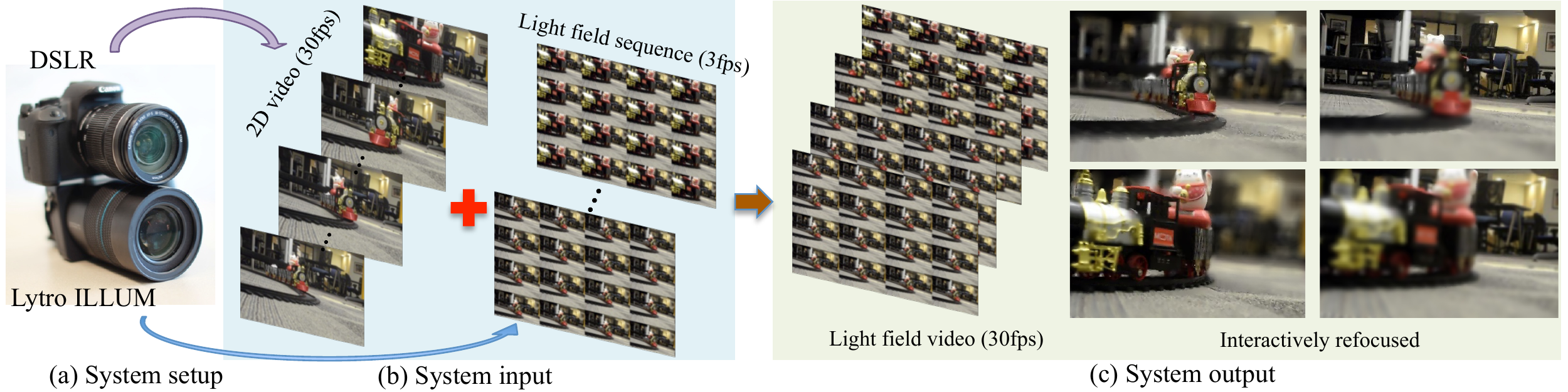}
   \caption{The setup and I/O of our system. (a) We attach an additional standard camera to a light field camera using a tripod screw, so they can be easily carried together.
   (b) The inputs consist of a standard $30$ fps video and a $3$ fps light field sequence.
   (c) Our system then generates a $30$ fps light field video, which can be used for a number of applications such as refocusing and changing viewpoints as the video plays.}
 \lblfig{teaser}
\end{teaserfigure}

\begin{abstract}
Light field cameras have many advantages over traditional cameras, as they allow the user to change various camera settings {\em after} capture.
However, capturing light fields requires a huge bandwidth to record the data: a modern light field camera can only take three images per second.  
This prevents current consumer light field cameras from capturing light field videos.
Temporal interpolation at such extreme scale (10x, from $3$ fps to $30$ fps) is infeasible as too much information will be entirely missing between adjacent frames.  
Instead, we develop a hybrid imaging system, adding another standard video camera to capture the temporal information.
Given a $3$ fps light field sequence and a standard $30$ fps 2D video, our system can then generate a full light field video at $30$ fps.
We adopt a learning-based approach, which can be decomposed into two steps: spatio-temporal flow estimation and appearance estimation.
The flow estimation propagates the angular information from the light field sequence to the 2D video, so we can warp input images to the target view.
The appearance estimation then combines these warped images to output the final pixels.
The whole process is trained end-to-end using convolutional neural networks.
Experimental results demonstrate that our algorithm outperforms current video interpolation methods, enabling consumer light field videography, and making applications such as refocusing and parallax view generation achievable on videos for the first time.
Code and data are available at 
\mbox{\footnotesize \url{https://cseweb.ucsd.edu/~viscomp/projects/LF/papers/SIG17/lfv/}}.
\end{abstract}


\setcopyright{acmcopyright}
\acmJournal{TOG}
\acmYear{2017}\acmVolume{36}\acmNumber{4}\acmArticle{133}\acmMonth{7} 
\acmDOI{http://dx.doi.org/10.1145/3072959.3073614}


\maketitle

\section{Introduction} \lblsec{intro}
Light field cameras have recently become available in the consumer market (e.g.\ Lytro), making applications such as post-shot photograph refocusing and viewpoint parallax possible.
The great promise of the light-field camera is that a lot of what used to be a pre-process (focus, aperture, etc) can now be a post-process -- you shoot first and then decide what you want later.  
The biggest advantage for this would be for shooting movies, as the director can experiment with changing focus within different parts of the video after it has been captured. 
It can also save the effort of focus pulling, i.e.\ manually shifting the focus plane to remain focused on a moving object within a shot.

However, recording scenes in both spatial and angular domains takes a significant amount of data, which limits the maximum data transfer (write) speed, given a limited bandwidth.
For example, the raw output image of the Lytro ILLUM camera is $5300\times7600$ pixels, which is nearly $20$ times the resolution of 1080p videos.
Assuming we are given the same bandwidth as a 1080p $60$ fps video, we can only record light field images at $3$ fps, which is exactly the rate of the continuous shooting mode for Lytro ILLUM. 
Although some modern film cameras such as the Red camera~\shortcite{red} can shoot at higher frame rate, they are usually very expensive, and cannot easily fit the budget for ``YouTube filmmakers'' -- artists making high-quality but low-budget work using consumer-grade equipment and non-traditional distribution channels. 
For this new democratized cinematography, a \$30-50k Red camera is usually out of reach.
This makes recording light field videos of normal activities impractical for these artists.

One possible solution is to use a cheaper consumer camera, and perform video interpolation.
That is, given the temporal sampling of the scene, interpolate between the input frames to recover the missing frames.
However, it is extremely hard to interpolate such a low frame-rate video.
In fact, some motion may be entirely missing between neighboring frames, making interpolation impossible, as shown in \reffig{aliasing}.

In this paper, we approach this problem in a fundamentally different way.
We develop a hybrid imaging system, combining the Lytro ILLUM camera with a video camera capable of capturing a standard $30$ fps video (\reffig{teaser}a).
The inputs to our system thus consist of a low frame-rate light field video and a standard 2D video (\reffig{teaser}b).\footnote{We downsample the video resolution to match the light field sub-aperture resolution. Therefore, the extra bandwidth required for recording the video is actually minimal (about $5\%$ more pixels), and in the future, the two cameras may be merged into one with some hardware modification.}
The $3$ fps light field video captures the angular information, while the $30$ fps video gathers the temporal information.
Our goal is to output a full light field video with all angular views at the standard video rate (\reffig{teaser}c left). 
This makes light field image applications achievable on videos for the first time using consumer cameras, such as digital refocusing and parallax view generation as the video is played (\reffig{teaser}c right). 

Given the sparse light field sequence and the 2D video, we propose a learning-based approach to combine these two sources of information into one.
We achieve this by propagating the angular information captured at the light field frames to the 2D-only frames.
For each target view in the missing light field frames, we solve a view synthesis problem.
We break down the synthesis process into two steps: a spatio-temporal flow estimation step and an appearance estimation step (\reffig{overview}).
The first step estimates the flows between both the 2D frames and the light field frames, and warps them to the target view accordingly (\refsec{algorithm:flow}, \reffig{flow_overview}a).
The second step then combines these warped images to output the final pixel color (\refsec{algorithm:color}, \reffig{flow_overview}b).

For both estimation steps, we adopt the recently popular convolutional neural network (CNN)~\cite{lecun1998gradient}.
An advantage of using CNN methods compared to traditional (i.e.\ non-learning) methods is that they can provide end-to-end training, so the generated results will usually be better than doing each step independently.
Another advantage is that CNNs are much faster than traditional methods.
For example, our flow estimation network is two orders of magnitude faster than the state-of-the-art optical flow method~\cite{revaud2015epicflow}.
Overall, our method takes less than one second to generate a novel view image, and after that, $0.06$ seconds for an additional view.
Also note that normally, training flow networks would require the ground truth flows, which are hard to obtain.
However, we show that we are able to train them by minimizing errors when the outputs are used to warp images, without utilizing ground truth (\reffig{flow_compare}).

To better visualize our results, we also develop an interactive user interface to play the resulting light field video, which allows the user to focus to any point as the video plays (\reffig{refocus}), track a given object and remain focused on it throughout the video (\reffig{focus_track}), change the (effective) aperture size to create different depths of field (\reffig{aperture}), and vary the viewpoints to provide a more lively viewing experience.
Examples of using our interface to achieve these effects can be found in the accompanying video. 

In summary, our contributions are:
\par \textbf{1)} We propose the first algorithm to generate a $30$ fps light field video using consumer cameras.
Experimental results demonstrate that this cannot be achieved by current state-of-the-art video interpolation and depth from video methods (Figs.~\ref{fig:temporal_compare}, ~\ref{fig:quant_compare} and~\ref{fig:angular_compare}).

\par \textbf{2)} We develop a CNN architecture to combine light field and 2D videos (Figs.~\ref{fig:overview} and~\ref{fig:flow_overview}).
In particular, we train a disparity CNN (\reffig{network_io}a) and an optical flow CNN (\reffig{network_io}b) without utilizing ground truth, and cascade them to combine the angular and the temporal information.

\begin{figure}[t!]
\vspace{-.1in}
  \centering
  \includegraphics[width=.95\linewidth]{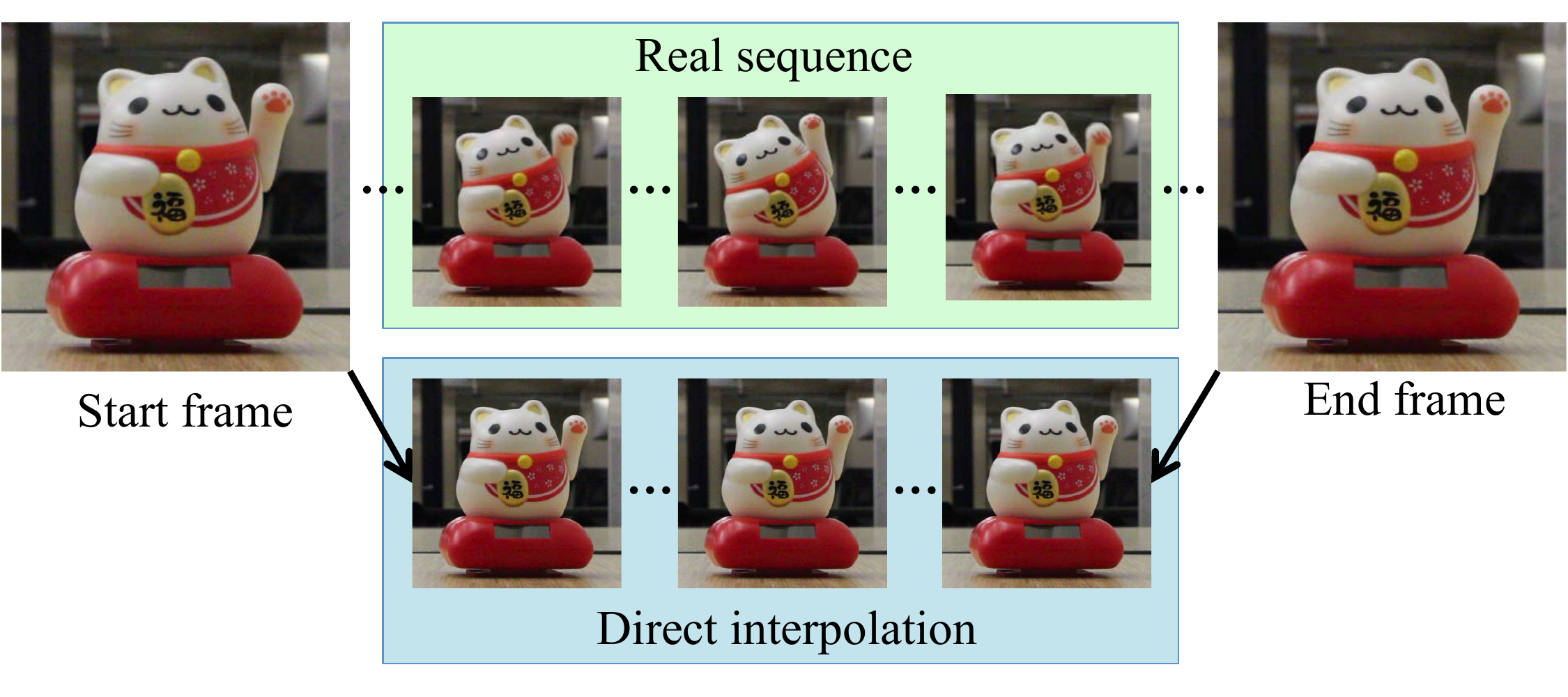}
  \vspace{-.1in}
  \caption{An example of temporal aliasing. Since the light field video is very low fps, directly interpolating between the sampled frames may lead to inaccurate results. In this example, the start frame and the end frame are the same, so all interpolated results are the same.}
  \lblfig{aliasing}
  \vspace{-.1in}
\end{figure}

\section{Related work} \lblsec{relatedwork}
\paragraph{Light Field Video}
The origin of light field videos goes back at least to~\cite{wilburn2002light}.
Nowadays, there are a few devices which can capture a light field video.
For example, some high-end models for RayTrix~\shortcite{RayTrix}, and the recent Lytro Cinema~\shortcite{Lytro}. 
However, these devices either have lower spatial or angular resolutions, or are very expensive, and are mainly targeted for research and not for ordinary users.
On the other hand, the consumer friendly Lytro ILLUM camera has an affordable price, but is not able to shoot light field videos.
In this work, we combine it with an additional 2D camera to interpolate a light field video.

\paragraph{Video Interpolation}
Many researchers have tried to upsample the frame rate of 2D videos by interpolating frames in the temporal domain~\cite{mahajan2009moving,baker2011database,liao2014automating,meyer2015phase}.
However, most existing approaches aim only for slow motion videos, taking a standard 30 fps video as input and trying to generate a very high fps video.
This approach, however, is in general not enough to deal with very low fps videos.
For a $30$ fps video, the motion between neighboring frames is usually very small, which makes interpolation much easier.
The same assumption does not hold for a $3$ fps video; in fact, some motion may be entirely missing between neighboring frames, making interpolation entirely impossible, as shown in \reffig{aliasing}. 
Moreover, a standard 2D video interpolation method will not produce a consistent light field video for applications like refocusing/viewpoint change, even if applied separately to each view. 
We show the advantage of our approach over existing 2D interpolation techniques in~\refsec{results}.

\paragraph{Light Field Super-resolution}
Since a light field has a limited resolution, many methods have been proposed to increase its spatial or angular resolution~\cite{bishop2009light,mitra2012light,cho2013modeling}.
Some require the input light fields to follow a specific format to reconstruct images at novel views~\cite{levin2010linear,marwah2013compressive,shi2014light}.
Wanner and Goldluecke~\shortcite{wanner2014variational} propose an optimization approach to synthesize new angular views.
Yoon et al.~\shortcite{yoon2015learning} apply convolutional neural networks (CNN) to perform spatial and angular super-resolution.
Zhang et al.~\shortcite{zhang2015light} propose a phase-based approach to reconstruct
light fields using a micro-baseline stereo pair.
To synthesize new views, Kalantari et al.~\shortcite{kalantari2016learning} break the problem into depth estimation and appearance estimation and train two sequential neural networks. 
Wu et al.~\shortcite{wu2017light} apply a CNN-based angular detail restoration on epipolar images to recover missing views.
However, none of these methods are designed for temporal upsampling, which is a completely different problem.

\paragraph{Hybrid Imaging System}
Combining cameras of two different types to complement each other has also been proposed before.
However, they are either used to increase the spatial resolution~\cite{sawhney2001hybrid,bhat2007using,boominathan2014improving,wang2016semi,wang2016light}, deblur the image~\cite{ben2003motion}, or create hyperspectral images~\cite{kawakami2011high,cao2011high}.
None of these methods has tried to increase the temporal resolution.

\paragraph{View Synthesis}
To synthesize novel views from a set of given images, many methods first estimate the depth and warp the input images to the target view using the obtained depth~\cite{eisemann2008floating,goesele2010ambient,chaurasia2013depth}.
The final synthesized image is then a combination of these warped images.
To generate a light field video, we also adopt a similar approach.
Inspired by~\cite{flynn2015deepstereo,kalantari2016learning,zhou2016view}, we use a learning-based approach to perform the geometry and appearance estimations.
However, instead of synthesizing images at new viewpoints, we perform image synthesis in the temporal domain.

\paragraph{2D to 3D Conversion}
Given a 2D video, there are many works that try to generate the corresponding depth sequence.
Konrad et al.~\shortcite{konrad20122d} propose a learning-based approach using millions of RGB+depth image pairs, and adopt the k nearest-neighbor (kNN) algorithm to obtain depths for a 2D query video.
They then further extend it to predict pixel-wise depth maps by learning a point mapping function from local image/video attributes~\cite{konrad2013learning}.
Karsch et al.~\shortcite{karsch2014depth} collect an RGBD dataset, and adopt non-parametric depth sampling to generate depths from a monoscopic video based on SIFT flows.
However, their method requires the entire training dataset to be available at runtime, which requires significant memory, and is very time consuming.
Besides, the above methods only generate depths, while our method can produce a light field sequence.

\section{Algorithm} \lblsec{algorithm}
\begin{figure*}[ht]
  \centering
  \includegraphics[width=\linewidth]{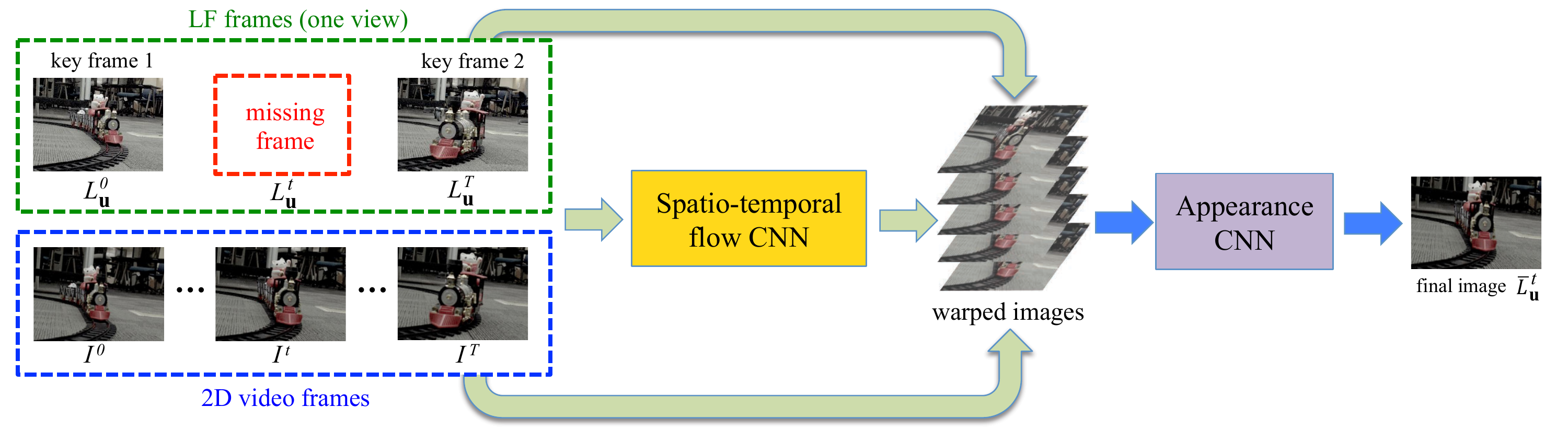}
  \vspace{-.2in}
  \caption{Overview of our system. Our system consists of two main parts: the spatio-temporal flow CNN and the appearance CNN. 
  The first CNN warps the input video frames and light field images to the target angular view. 
  The second CNN then combines all these warped views to generate a final image. 
  Note that only one view in the light fields is shown here for simplicity.
  For now, we assume the central view matches the 2D video view for easier explanation.}
  \lblfig{overview}
\end{figure*}

\begin{figure*}[ht!]
  \centering
  \includegraphics[width=\linewidth]{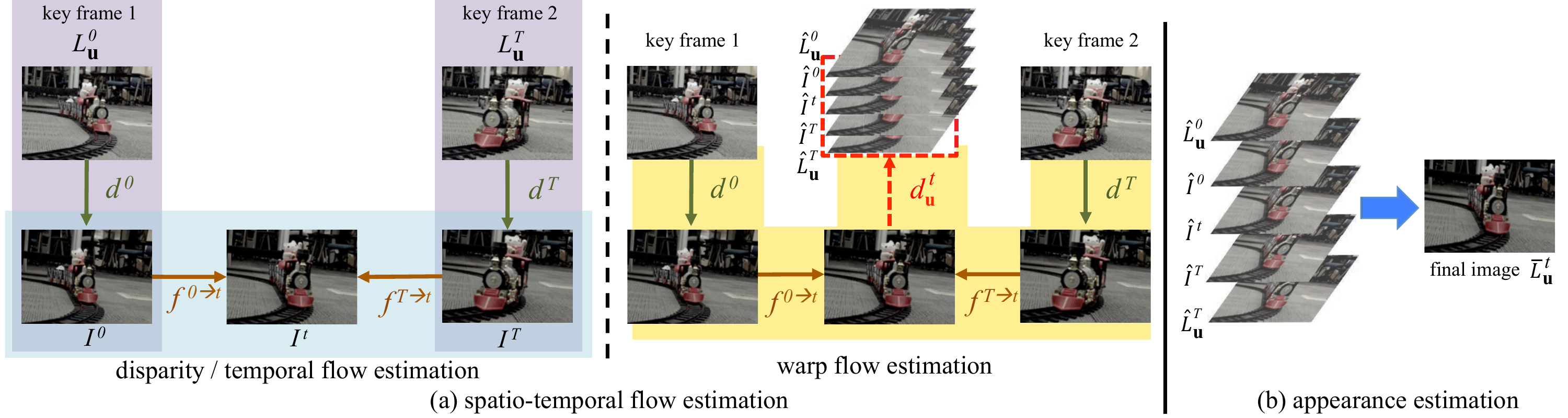}
  \vspace{-.2in}
  \caption{(a) Overall architecture for our spatio-temporal flow estimation network. The system contains three components: disparity estimation, temporal flow estimation, and warp flow estimation. 
  The first (left) part shows the first two components.
  First, we estimate the disparity at the key frames, as well as the temporal flow between the key frame and the current frame in the 2D video.
  Afterwards, the estimated flows are concatenated to generate five flows that warp the five input images to the target view.
  These five images are the two light field angular views at the previous and next keyframes, the two corresponding 2D video frames, and the current 2D frame.
  (b) Finally, these warped images are put into the appearance estimation network to output the final image.}  
  \lblfig{flow_overview}
\end{figure*}

Given a standard ($30$ fps) 2D video and a very low speed light field video (e.g.\ $3$ fps), our goal is to generate a full light field video.
Since some of the 2D frames have corresponding light fields (denoted as \textit{key frames}) while others do not, our main idea is to propagate this multi-view information from these keyframes to the in-between frames.
As illustrated in~\reffig{overview}, the overall architecture of our system contains two main parts: spatio-temporal flow estimation CNN and appearance estimation CNN.
The spatio-temporal CNN warps the input images from the 2D video and the light field images to the target angular view; the appearance CNN then combines all the warped images to generate a final image.
We chose to train neural networks over traditional methods for two reasons. 
First, using neural networks enables end-to-end training, which typically yields better performance than manually designing each component independently.
Second, neural networks usually run $10\sim100$ times faster than traditional methods.

Note that the light field camera and the video camera will have slightly different viewpoints.
For simplicity, we first assume the central view of the light field camera coincides with the 2D video viewpoint.
In practice we estimate another flow between the two images to calibrate for that, with more details given in~\refsec{implementation}. 
We also discuss how to handle different color responses between the two cameras in~\refsec{implementation}.
To make things simpler, we downsample the DSLR resolution to match the Lytro resolution, since spatial super-resolution is not the focus of this paper.

We denote the 2D video frames by $I^t$ and light field sequences by $L^t$, where $t = 1,2,3...$ is the frame index.
Let $L^0$ and $L^T$ be two neighboring keyframes.
Our problem can then be formulated as: given $(L^0,L^T)$ and $\{I^0,I^1,...,I^{T-1},I^T\}$, estimate $\{L^1,...,L^{T-1}\}$. 
We only compute intermediate frames between two adjacent keyframes $(L^0,L^T)$, and later concatenate all the interpolated results to produce the full video.

\subsection{Spatio-temporal flow estimation network} \lblsec{algorithm:flow}
The spatio-temporal flow network can be divided into three components: disparity estimation, temporal flow estimation, and warp flow estimation, as shown in Fig.~\ref{fig:flow_overview}. 
The first component estimates the disparities at the key frames.
The second component (temporal flow estimation) then computes how we can propagate information between the 2D frames. 
Finally, the third component (warp flow estimation) utilizes the results of the previous two components and propagates disparities to all the 2D frames.
We first train each component independently and then end-to-end, along with the appearance estimation network.
The inputs and outputs of each component are shown in \reffig{network_io}, and described below.  
The actual network architectures will be discussed in \refsec{implementation}.
\begin{figure}[ht!]
  \centering
  \includegraphics[width=\linewidth]{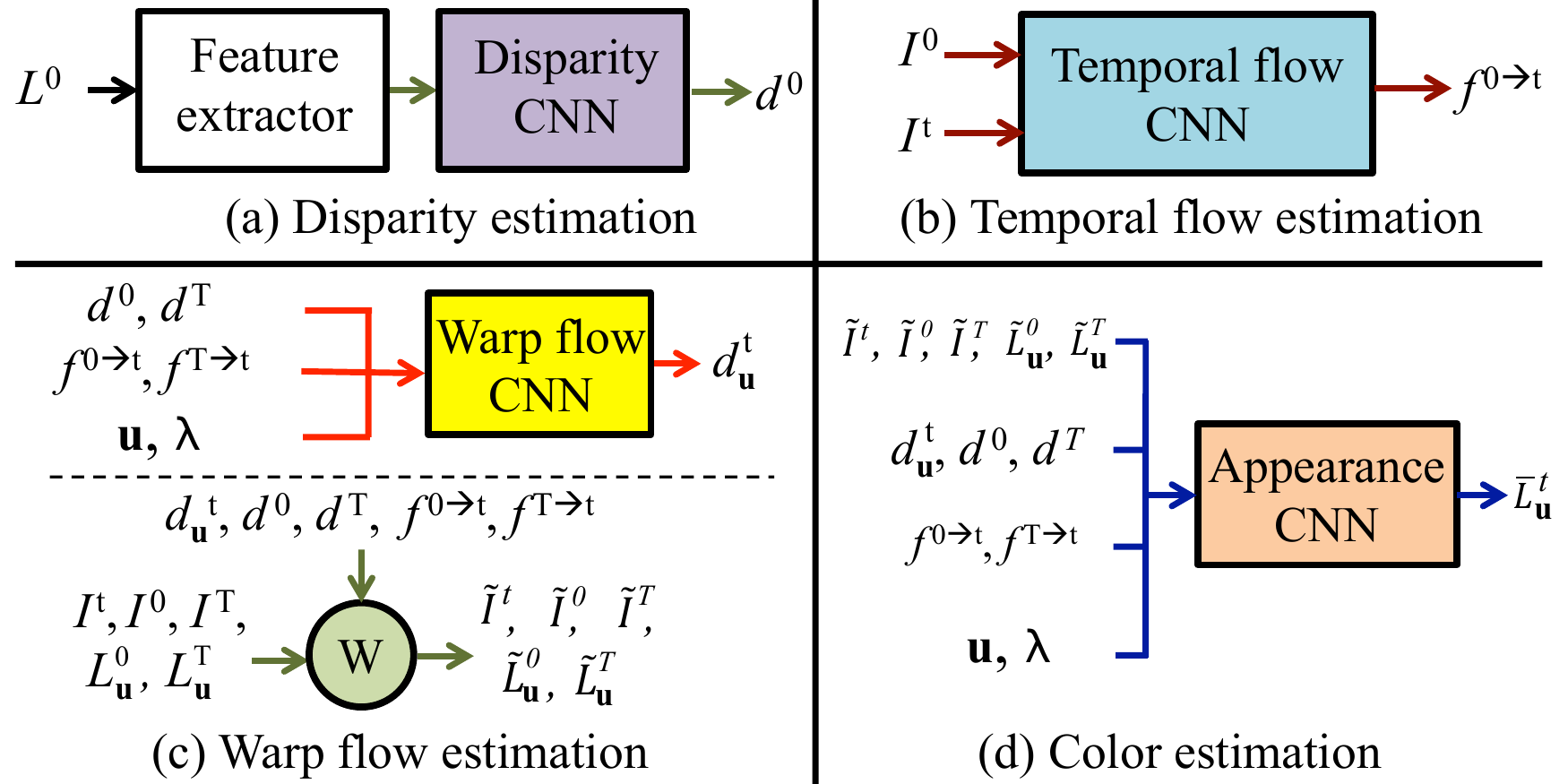}
  \caption{The I/O of each component in our networks.
  (a) Given the light field at one keyframe $0$, the disparity CNN estimates the central view disparity.
  (b) Given two frames $0$ and $t$ in the 2D video, the temporal flow CNN generates the optical flow between them.
  (c) Given the estimated disparities and flows, the warp flow CNN computes the disparity at target view $L_\textbf{u}^t$ (upper half). 
  This disparity is then concatenated with the other disparities/flows to generate five flows, which are used to warp the five images to the target view (lower half).
  (d) Finally, the warped images along with disparities/flows are stacked together and fed into the appearance CNN to output the final image.}  
  \lblfig{network_io}
  \vspace{-.1in}
\end{figure}

\paragraph{Disparity estimation}
Given the light field views of one frame (denoted as $L^0$ or $L^T$), we first try to estimate the disparity of this frame at the central view.
Our method is similar to the disparity CNN proposed by Kalantari et al.~\shortcite{kalantari2016learning}, which is inspired by~\cite{flynn2015deepstereo}.
We briefly introduce the idea here.
For each disparity level, we first shift all the views by the corresponding amounts. 
Ideally, if we shift them by the correct amount, all views will overlap and their mean should be a sharp (in focus) image and their variance should be zero. 
Therefore, the means and variances of the shifted views are extracted to form a $h\times w\times 2$ feature map, where $(h,w)$ is the image size. 
This process is repeated for $n$ different disparity levels, and the features are all concatenated together to form a $h\times w\times 2n$ feature map. 
This feature is then input to a 4-layer fully convolutional network to generate the depth map.

The only difference of our network from~\cite{kalantari2016learning} is that we have all $64$ light field views instead of just $4$ corner views,\footnote{The angular resolution of the Lytro ILLUM is $14\times 14$. However, we only use the central $8 \times 8$ views as the views on the border are dark.}
so we shift all the $64$ views accordingly, and take the mean and variance of the shifted views.
These shifted images are then put into a CNN to output the disparity $d(x,y)$ at the key frame.
In short, the inputs to this network are the $64$ views of key frame $L^0$, and the output is the disparity $d^0$ (\reffig{network_io}a).
The same process is performed for the other key frame $L^T$ to obtain $d^T$.

To train this network, the usual way is to minimize the difference between the output disparity and the ground truth.
However, ground truth depths are hard to obtain.
Instead, we try to minimize the loss when the output depth is used to warp other views to the central view, which is similar to~\cite{kalantari2016learning}.
In particular, if we assume a Lambertian surface, the relationship between the central view $L(x,y,0,0)$ and the other views $L(x,y,u,v)$ can be modeled as
\begin{equation}
\begin{split}
L^0(x,y,0,0) &= L^0(x+u\cdot d^0(x,y),y+v\cdot d^0(x,y), u, v) \\
&= L^0(\textbf{x}+\textbf{u}\cdot d^0(\textbf{x}), \textbf{u})
\end{split}
\lbleq{disparity}
\end{equation}
where $\textbf{x}=(x,y)$ is the spatial coordinate and $\textbf{u}=(u,v)$ is the angular coordinate.
The loss we try to minimize is then the reconstruction difference (measured by Euclidean distance) between the two sides of (\ref{eq:disparity}), summed over all viewpoints,
\begin{equation}
E_d(\textbf{x}) = \sum_{\textbf{u}} || L^0(\textbf{x}, \textbf{0}) - L^0(\textbf{x}+\textbf{u}\cdot d^0(\textbf{x}), \textbf{u})||^2
\lbleq{disparity_cost}
\end{equation}

\begin{figure}[t!]
  \centering
  \includegraphics[width=.95\linewidth]{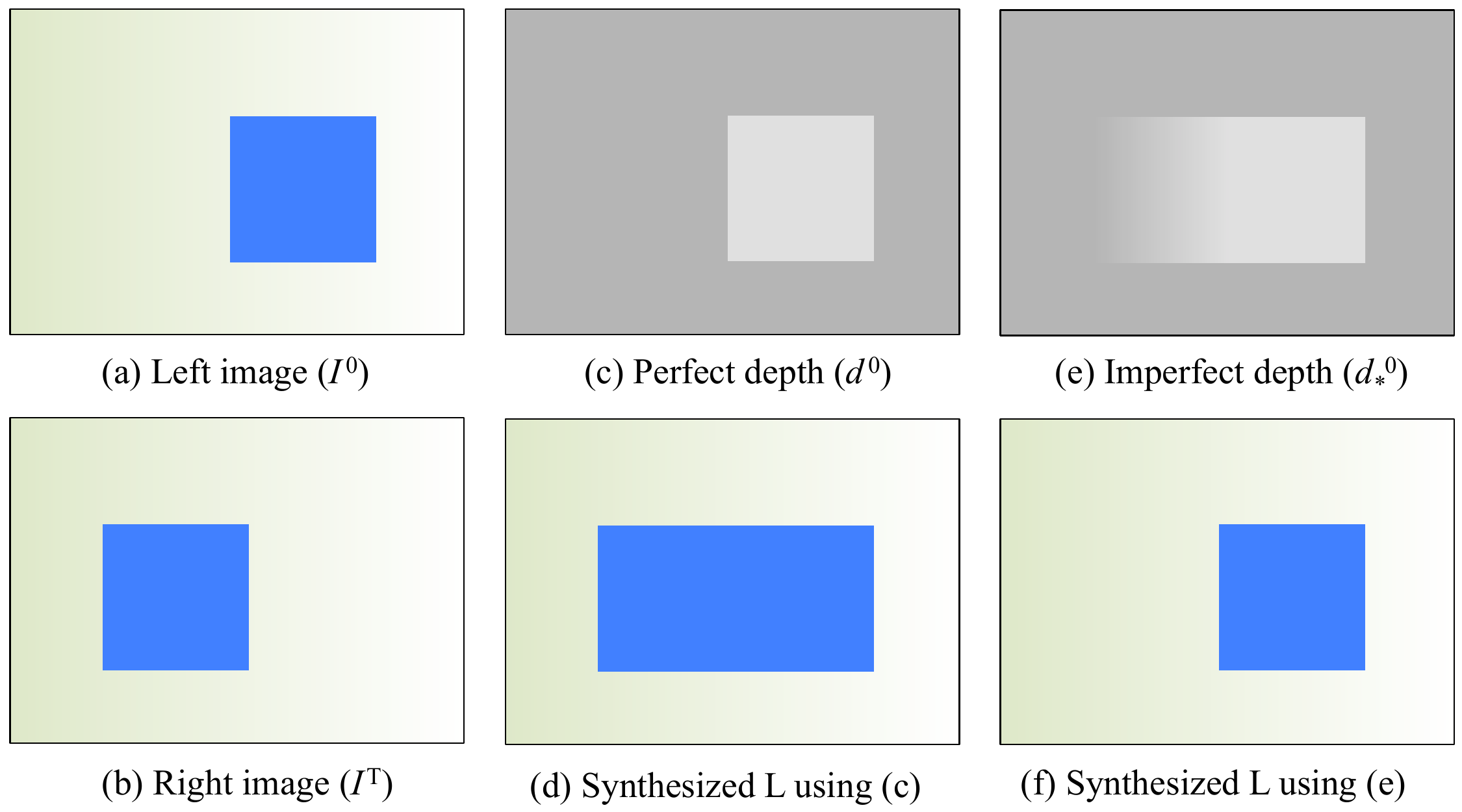}
  \vspace{-.1in}
  \caption{An example illustrating that perfect depths are not necessarily good for synthesizing views. (a)(b) To make things simple, we just consider a stereo pair input. 
  (c) A perfect depth for viewpoint $0$. Brighter colors represent nearer depths. 
  (d) Using $d^0$ to synthesize $I^0$ from $I^T$. Note that although the depth is correct, it tries to borrow pixels that are already occluded, and get pixels from the occluder instead. 
  (e) An imperfect depth, which will actually synthesize a better view as shown in (f).}
  \label{fig:occlusion}
  \vspace{-.1in}
\end{figure}

Note that by computing the loss function this way, although we do not have the ground truth depth maps, we are still able to optimize for a functional depth map for the purpose of image synthesis. 
Also, although we can use methods other than CNN to compute the disparities directly as well, we found that images synthesized by these methods often produce visible artifacts, especially around occlusions. 
This is because they are not specifically designed for view synthesis. 
A schematic example is shown in Fig.~\ref{fig:occlusion}. 
Note that in this case, a ``perfect'' depth is actually not able to reconstruct the view around the occlusion boundaries. 
Instead, a fuzzy depth map will generate a more visually pleasing result.
The advantage of using learning-based methods is that although we do not explicitly model occlusions, the network will still try to find some way to handle them.

\paragraph{Temporal flow estimation}
We now try to estimate the optical flow between the key frames and the in-between frames in the 2D video.
We do this by estimating flows between every pair of neighboring frames in the video, and cascade the flows to get the flow between the key frame and the other frames.
Without loss of generality, we consider the case where we estimate the direct flows between two frames $I^0$ and $I^t$ below.
The inputs to this network are these two frames.
The output is the flow $f^{0\rightarrow t}$ that warps $I^0$ to $I^t$ (\reffig{network_io}b).

Unlike the disparity estimation, this process is much harder for two reasons.
First, disparity is a 1D problem, while the general flow is 2D.
Second, the pixel displacement in a temporal flow is usually much larger than the case in light fields, so the search range needs to be much larger as well.
Based on these conditions, if we use the same architecture as in the previous disparity network, the feature map we extract will have an intractable size. 
For example, in \cite{kalantari2016learning} each shift amount differs by about 0.4 pixels. 
If we want to retain the same precision for flows up to $100$ pixels, our feature map will be $(200/0.4)^2 \times 2 = 500,000$ dimensional, which is clearly impractical.

\begin{figure}[t!]
  \centering
  \begin{subfigure}[b]{\linewidth}
  \centering
  \includegraphics[width=.95\linewidth]{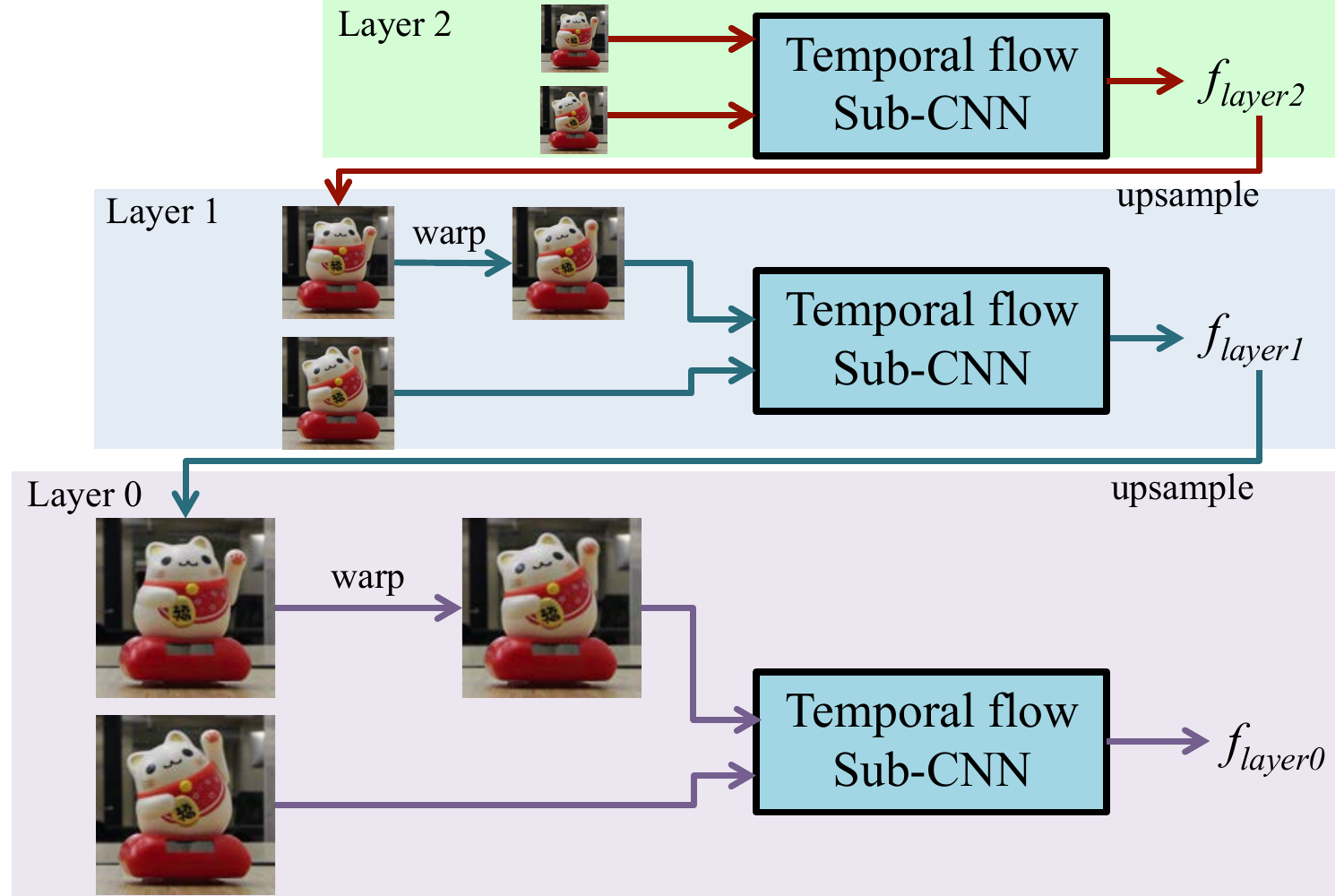}  
  \subcaption{Hierarchical architecture of our temporal flow network.}
  \end{subfigure} \\
  \begin{subfigure}[b]{\linewidth}
  \centering
  \includegraphics[width=.95\linewidth]{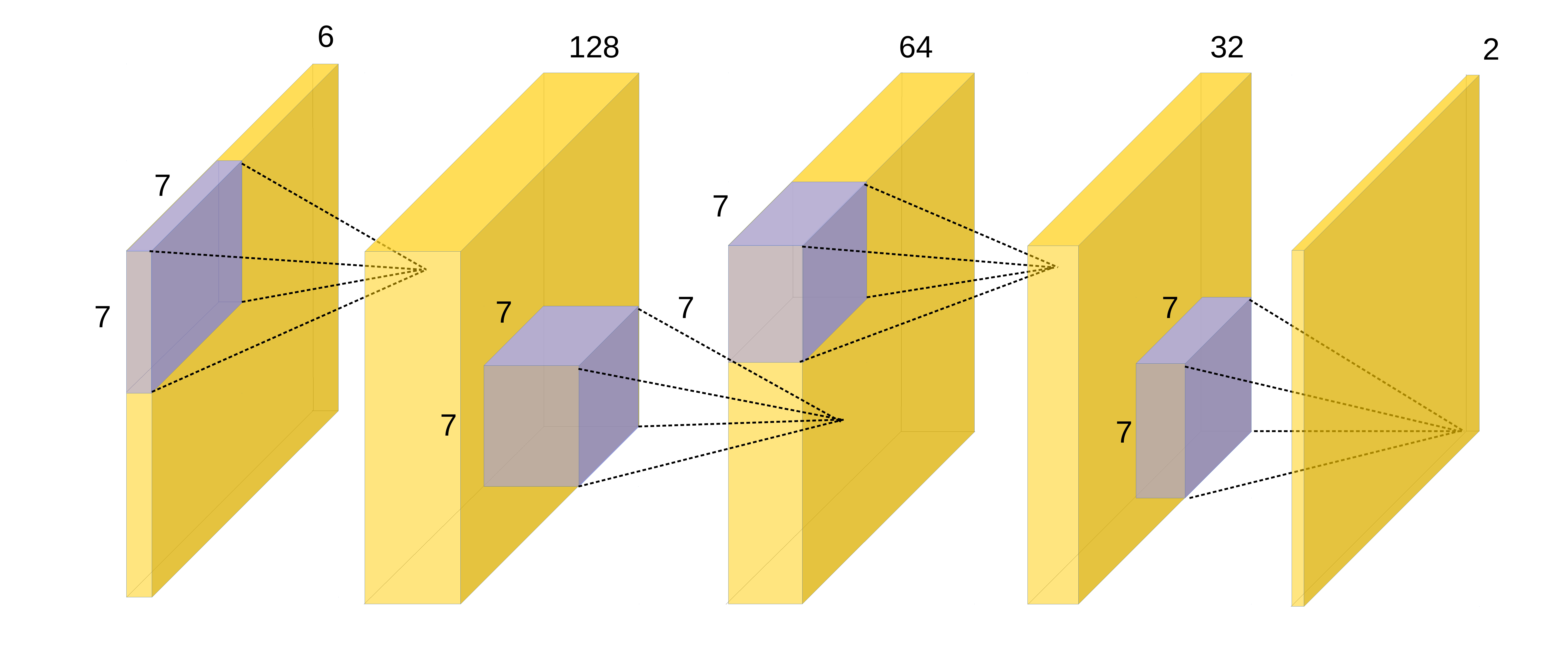}  
  \subcaption{Architecture of the sub-CNNs in (a).}
  \end{subfigure}
  \vspace{-.2in}
  \caption{Our temporal flow CNN. (a) We adopt a hierarchical approach. 
  For the top (coarsest) layer, the flow estimation works normally.
  For subsequent layers, we first warp the input using the flow estimated from the previous layer, then perform flow estimation on the warped images.
   This process is repeated until the flow in the finest level is estimated. (b) Our sub-CNN contains 4 conv layers. Each convolution except for the last one is followed by a rectified linear unit (ReLU).}
  \label{fig:temporal}
  \vspace{-.2in}
\end{figure}

To resolve this, we adopt a hierarchical approach instead (Fig.~\ref{fig:temporal}).
For the two input frames $I^0$ and $I^t$, we build a Gaussian pyramid for each of them.
Optical flow is then estimated at the coarsest level, and propagated to the finer level below.
This process is repeated until we reach the finest level to get the final flow.

Similar to the case of disparity estimation, since ground truth flows are not available, we try to optimize by using the output flow to warp images.
More specifically, the relationship between $I^0$ and $I^t$ can be written as
\begin{equation}
I^t(\textbf{x}) = I^0(\textbf{x}+f^{0\rightarrow t}(\textbf{x}))
\label{eq:optical_flow}
\end{equation}
The loss we try to minimize is the Euclidean distance between the two sides of (\ref{eq:optical_flow}),
\begin{equation}
E_f(\textbf{x}) = || I^t(\textbf{x}) - I^0(\textbf{x}+f^{0\rightarrow t}(\textbf{x}))||^2
\lbleq{flow_cost}
\end{equation}
Note that this is in contrast to other optical flow CNN training procedures such as FlowNet, which requires the ground truth flows.

\paragraph{Warp flow estimation}
This step warps the five images shown in Fig.~\ref{fig:overview} to the target image $L_\textbf{u}^t$ (missing frame).
The five images are: the current 2D frame $I^t$, the two 2D frames $I^0$, $I^T$ and two target views $L_\textbf{u}^0$, $L_\textbf{u}^T$ at the keyframes.
Note that $I^t$ is usually closer to our final output and often contributes the most, since the angular motion is usually smaller than the temporal motion.
However, in some cases the other images will have more impact on the result.
For example, when part of the scene is static, $L_\textbf{u}^0$ will just remain the same throughout the video, so $L_\textbf{u}^0$ (or $L_\textbf{u}^T$) will be closer to $L_\textbf{u}^t$ than $I^t$ is to $L_\textbf{u}^t$.

To generate the warp flows that warp these images to the target view, we first estimate the disparity $d_{\textbf{u}}^t$ at the target view $\textbf{u}$ at the current frame $t$ (\reffig{network_io}c up).
We do this by utilizing the disparities obtained from key frames $0$ and $T$ to generate the central view disparity $d^t$ first.
To utilize the disparity at key frame $0$, we can first ``borrow'' its disparity in the same way we borrow its color pixels,
\begin{equation}
d^t(\textbf{x}) = d^0(\textbf{x} +f^{0\rightarrow t}(\textbf{x}))
\end{equation}
Similarly, we can also borrow the disparity from key frame $T$,
\begin{equation}
d^t(\textbf{x}) = d^T(\textbf{x} +f^{T\rightarrow t}(\textbf{x}))
\end{equation}
The final disparity should be somewhere between these two ``borrowed'' disparities.
Intuitively, when we are closer to frame $0$, the current disparity should be closer to $d^0$, so the weight for it should be higher, and vice versa.
We thus add the temporal position $\lambda$ as an input to the CNN, indicating this ``closeness''
\begin{equation}
\lambda = t/T
\end{equation}

However, since we have no idea how the object depth changes between these two timeframes, there is no simple way to combine the two disparities.
Moreover, what we have now is the disparity which warps the target view $\textbf{u}$ to the central view (forward mapping), while we are actually interested in warping the central view to the target view (backward mapping).
This is in general very complex and no easy solution exists.
Therefore, we try to learn this transformation using a neural network.
We thus put the two borrowed disparities and $\lambda$ into a CNN to output the final disparity.
The output of the network is $d_{\textbf{u}}^t$ which satisfies
\begin{equation}
L^t(\textbf{x},\textbf{u}) = I^t(\textbf{x}-\textbf{u}\cdot d_{\textbf{u}}^t(\textbf{x}))
\label{eq:disparity_backward}
\end{equation}
The loss function of this network is then the Euclidean distance between the two images,
\begin{equation}
E_w(\textbf{x}) = || L^t(\textbf{x},\textbf{u}) - I^t(\textbf{x}-\textbf{u}\cdot d_{\textbf{u}}^t(\textbf{x}))||^2
\lbleq{warp_cost}
\end{equation}
Note that $L^t(\textbf{x},\textbf{u})$ is only available in training data, which are collected using the method described in Sec.~\ref{sec:implementation:train}.

\paragraph{Warping}
After we obtain the disparity $d_{\textbf{u}}^t$, we are now finished with the estimation part of this step (\reffig{network_io}c up), and can move on to the warping part (\reffig{network_io}c bottom).
Note that this part is entirely procedural and does not involve any CNNs.
We warp all neighboring images to the missing target view by cascading the flows. 
For example, let $\textbf{y}\equiv \textbf{x}-\textbf{u}\cdot d_{\textbf{u}}^t(\textbf{x})$. 
Then since $I^t(\textbf{y}) = I^0(\textbf{y}+f^{0\rightarrow t}(\textbf{y}))$ from (\ref{eq:optical_flow}), we can rewrite (\ref{eq:disparity_backward}) as
\begin{equation}
\begin{split}
L^t(\textbf{x}, \textbf{u}) &= I^t(\textbf{y}) = I^0(\textbf{y}+f^{0\rightarrow t}(\textbf{y})) \\
&= I^0(\textbf{x}-\textbf{u}\cdot d_{\textbf{u}}^t(\textbf{x}) +f^{0\rightarrow t}(\textbf{x}-\textbf{u}\cdot d_{\textbf{u}}^t(\textbf{x})))
\end{split}
\label{eq:I0_warp}
\end{equation}
which warps frame $0$ of the 2D video to the current target view.

The above example shows how we can warp one input frame, namely the lower left input image $I^0$ in Fig.~\ref{fig:overview} to $L^t$.
We denote this warped image as $\widetilde{I^0}$, where the tilde symbol indicates warped images.
In the same way, we can also warp the four other input images in Fig.~\ref{fig:overview}, generating five warped images $\widetilde{I}^t$,$\widetilde{I}^0$, $\widetilde{I}^T$, $\widetilde{L}_\textbf{u}^0$, and $\widetilde{L}_\textbf{u}^T$. 
The warp flows for all the five images are discussed in Appendix~\ref{sec:concate}.
These five images are then inputs to the next (appearance estimation) neural network to estimate $L^t(\textbf{x},\textbf{u})$.

\subsection{Appearance estimation network} \lblsec{algorithm:color}
After we warp all the images to the target view, we need to find a way to combine all these images to generate the final output.
Existing approaches usually use a simple weighted average to linearly combine them.
However, these simple approaches are usually not sufficient to handle cases in which the warped images differ significantly.
Instead, we train another network to combine them, inspired by~\cite{kalantari2016learning}.
We stack all five warped images together, along with the disparities and flows we estimated, the target view position $\textbf{u}$, and the temporal position $\lambda$, and put them into a network to generate the final image (\reffig{network_io}d).
The disparities and flows should be useful when detecting occlusion boundaries and choosing different images we want to use.
The angular and the temporal positions indicate which images should be weighted more when combining the warped images.
Note that unlike~\cite{kalantari2016learning}, where the input images differ only in the angular domain, here our input images are different in both angular and temporal domains, so the weighting mechanism is even more complex.
The final output $\bar{L}^t(\textbf{x},\textbf{u})$ of the network is the target image.
The loss of this network is thus
\begin{equation}
E_c(\textbf{x}) = || \bar{L}^t(\textbf{x},\textbf{u}) - L^t(\textbf{x},\textbf{u})||^2
\lbleq{overall_cost}
\end{equation}
where $L^t(\textbf{x},\textbf{u})$ is available in the training data.
The same procedure is performed for each target view and each time frame to generate the corresponding images.

\section{Implementation} \lblsec{implementation}
We first explain our network architectures in \refsec{implementation:network}, then describe how we train and test our models in \refsec{implementation:train} and \refsec{implementation:test}.

\subsection{Network architecture} \lblsec{implementation:network}
For our disparity CNN, the architecture is similar to \cite{kalantari2016learning}.
For the temporal flow CNN, we adopt a hierarchical approach to estimate the optical flow (Fig.~\ref{fig:temporal}).
In particular, we first build a pyramid for the two input images.
After the flow in the coarsest level is estimated, it is upsampled and used to warp the finer image below.
The warped images are then inputs to the network in the next level.
This process is repeated until the finest flow is estimated (Fig.~\ref{fig:temporal}a).
For each layer in the pyramid, the network architectures are the same and are shown in Fig.~\ref{fig:temporal}b.
For the warp flow CNN and the appearance CNN, the network architectures are similar to the  sub-CNN in the temporal flow CNN.

\begin{figure}[t!]
  \centering
  \includegraphics[width=.95\linewidth]{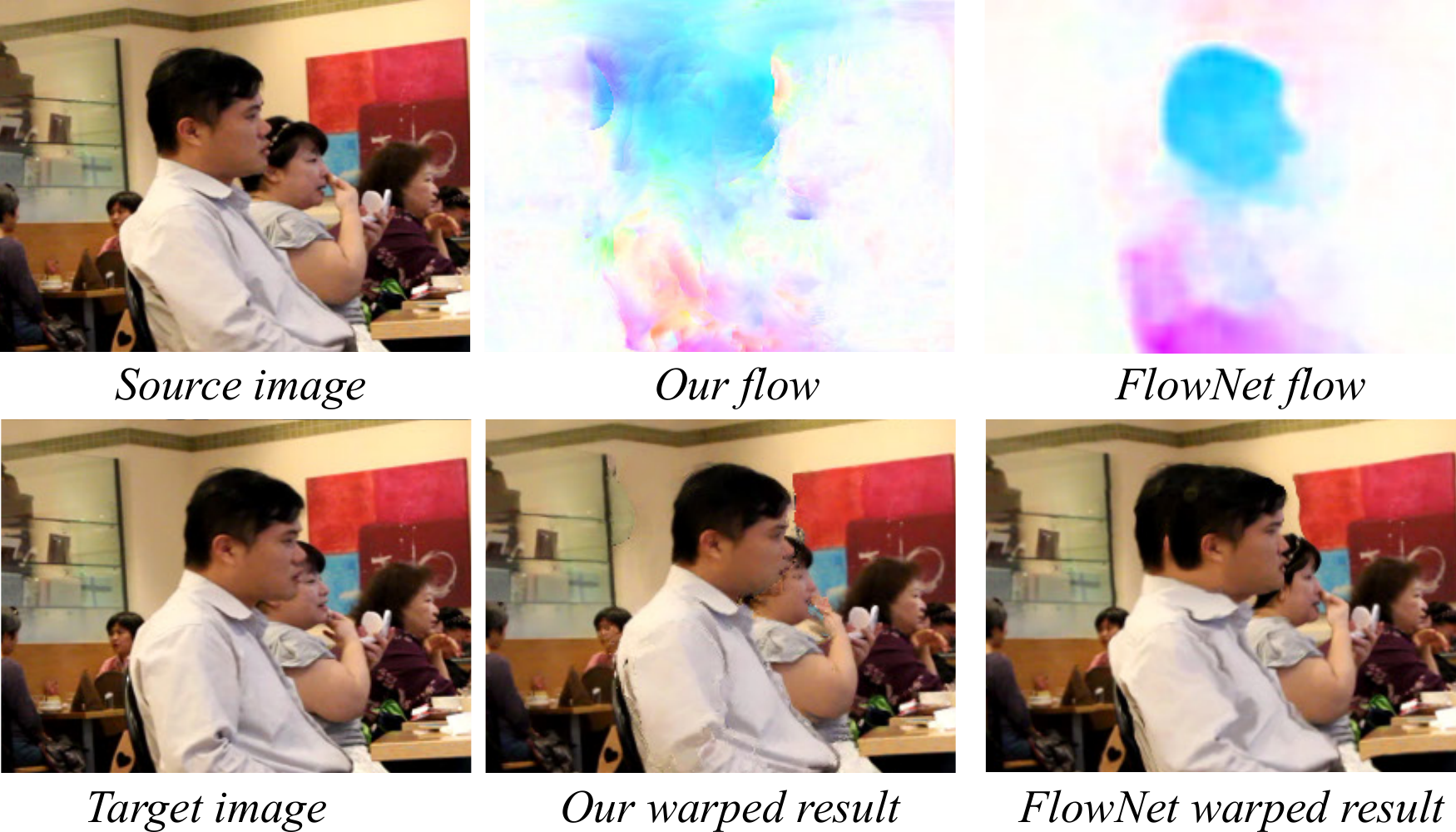}
  \vspace{-.1in}
  \caption{Comparison between our flow and flow by FlowNet.
  Although our flow is not as good as the result by FlowNet, it generates fewer artifacts when used to warp images, especially around the occlusion boundaries.
  Note that the region around the man's head, which was occluded in the source image, cannot be reproduced well by FlowNet, since it tries to borrow pixels that are already occluded, similar to the case in \reffig{occlusion}.}
  \label{fig:flow_compare}
    \vspace{-.1in}
\end{figure}

Finally, note that we train our network to generate better final images with an image reconstruction loss function, which is different from previous CNN-based optical flow methods (e.g.\ FlowNet~\cite{fischer2015flownet}). 
In their cases, the ground truth optical flow is available for networks to directly minimize the flow error.
However, we found our approach usually generates more visually pleasing results when warping images, since we are explicitly minimizing the color errors.
In particular, in the case of occlusions, the correct flow will actually try to borrow a pixel that is occluded by some other object, resulting in visible artifacts (Fig.~\ref{fig:flow_compare}).
Although our flow often looks worse, we are free of this problem by minimizing color errors instead of flow errors.

\subsection{Training and Evaluation}  \lblsec{implementation:train}
We describe how we obtain the ground truth and some details of training our model in this subsection.
Upon publication, the source code, the model and the datasets will be released, which will be a useful resource for further research on light field videos.

\paragraph{Ground truth data} 
To collect training data for our system, we shoot about $120$ diverse scenes with slow camera motion or slowly varying object motion using the Lytro ILLUM camera.
By collecting data this way, we are able to capture enough temporal information from this ``slow motion'' light field video as ground truth. 
Each sequence contains about $10$ frames.
For each recorded video clip, we treat the first frame and the last frame as keyframes, and the central views of light field images as 2D video frames.
We then generate other angular views for the $8$ intermediate light field frames via our approach, and compare the generated output with the ground truth data we captured. 
Example training and test scenes are shown in the supplementary video.

\paragraph{Training details} 
We now train a network to estimate the light fields for each of the in-between frames in our captured training data.
For each training example, we randomly crop the image to extract patches of size $320\times 320$.
In this way, we have effectively more than 10,000,000 patches for training.
We also perform standard data augmentation methods such as randomly scaling the images between $[0.9, 1.1]$ and swapping color channels.
We initialized the network weights using the MSRA filler~\cite{he2015delving} and trained our system using the ADAM solver~\cite{kingma2014adam}, with $\beta_1 = 0.9$, $\beta_2 = 0.999$, and a learning rate of $0.0001$.
At every iteration of training, we update the weights of our network using the gradient descent approach. 
Since each operation of our network is differentiable (convolutions, warpings, etc), computing the gradients can be easily done using the chain rule.

To ease the training difficulty, we first perform stage-wise training of each component in our system, and then perform joint training on all components. 
This is a common trick to train complex networks~\cite{hinton2006reducing}.

For the stage-wise training, the losses for the three components of the spatio-temporal flow network are stated in (\ref{eq:disparity_cost}), (\ref{eq:flow_cost}) and (\ref{eq:warp_cost}), respectively.
After this network is trained, we fix its weights to generate the inputs to the appearance estimation network, and train the appearance estimation network with the loss described in (\ref{eq:overall_cost}).

For joint training, we train both the spatio-temporal flow CNN and the appearance CNN jointly to optimize the cost function defined in \refeqshort{overall_cost}. 
This is basically the same as training the appearance CNN only, but allowing the weights in the spatio-temporal flow CNN to change as well.
We initialize the weights of each network from what we obtained from stage-wise training. 
To reduce the memory consumption of joint training, we use a few angular views instead of all the $64$ views in the disparity CNN. 
More training details can be found in Appendix~\ref{sec:training}.

\paragraph{Quantitative Evaluation} 
To evaluate our method quantitatively, we shot $30$ more scenes using the same method (i.e.\ slow camera motion and object motion) that we used to acquire the training data.  
We treat this as the test dataset for quantitative evaluation, and compare the image reconstruction error of our method against other previous approaches in \refsec{results:quant}. 

\begin{figure}[t!]
  \centering
    \includegraphics[width=.95\linewidth]{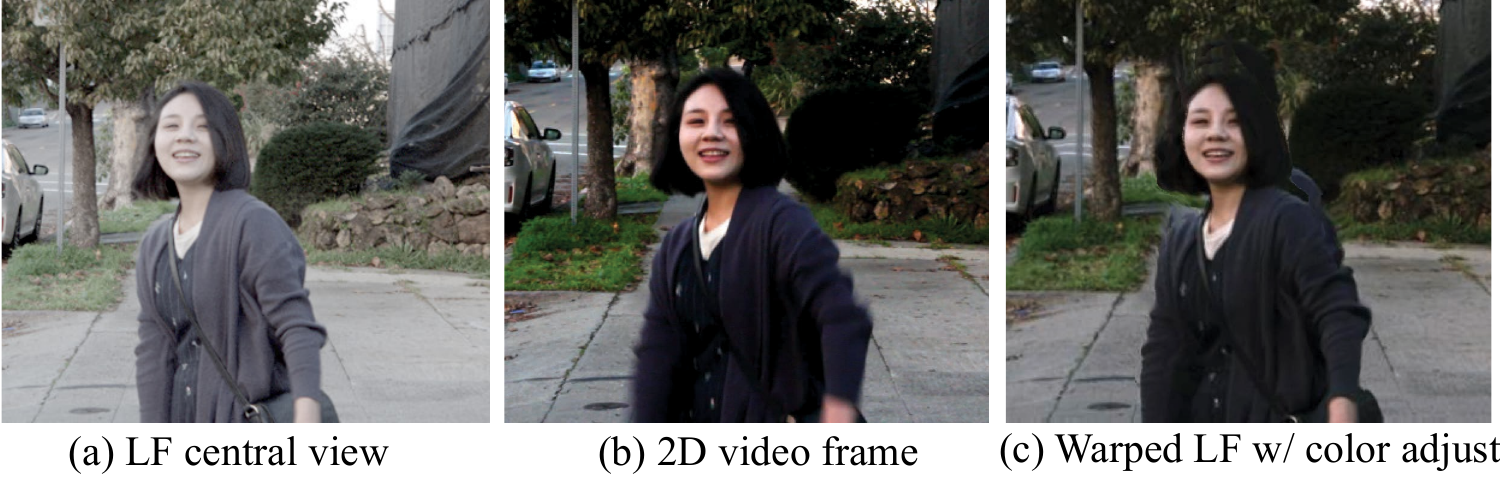}
    \vspace{-.1in}
  \caption{An example of NRDC calibration. We warp the light field view (a) to the video view (b) and perform color calibration using NRDC to obtain the warped result (c).}
  \lblfig{nrdc}
    \vspace{-.1in}
\end{figure}

\subsection{Testing in real-world scenarios} \lblsec{implementation:test}
For real-world scenarios where the ``ideal'' central view of light field images is inaccessible, we build a prototype in which a DSLR is connected with the Lytro ILLUM via a tripod screw adapter (Fig.~\ref{fig:teaser}a). 
The DSLR records a standard $30$ fps 2D video (rotated by $180$ degrees), while the Lytro camera captures a $3$ fps light field sequence, both of which are used as inputs to our network.

\paragraph{Camera calibration}
Since the DSLR view does not exactly match the central view of the Lytro camera in terms of viewpoint and color statistics, we first calibrate them by estimating a flow between the two views using non rigid dense correspondence (NRDC)~\cite{hacohen2011non}.
We chose NRDC due to its robustness to color differences between two cameras.
Furthermore, it can output the color transfer function between the two images, so that they can look identical. 
Figure~\ref{fig:nrdc} shows example results before and after the calibration.
Finally, to reduce blur in the DSLR image, we always use an aperture of $f/11$ or smaller for the DSLR.

\paragraph{Camera synchronization}
Next, to synchronize the images taken with the two cameras, we use the shutter sound the Lytro camera generates when taking images, which is recorded in the video.
Typically, there will be roughly $10$ to $15$ 2D frames between the Lytro camera shots, since the continuous shooting mode does not take images very steadily.
Using the sound to dynamically adjust to that gives us better matches than relying on a fixed number.

\paragraph{Testing details}
After these two calibrations, the same procedure follows as in the training process except the following differences.
First, we downsample the DSLR image to match the Lytro camera resolution.
Second, we use all the $64$ views in the keyframes rather than randomly sampling $4$ views to compute the disparity maps.
Finally, we now need to generate all the angular views (including the central view) for each 2D video frame.

\section{Results} \lblsec{results}
Below we compare our results both quantitatively (\refsec{results:quant}) and qualitatively (\refsec{results:qual}) to other methods, and show applications (\refsec{results:app}) and discuss limitations (\refsec{results:limit}) of our system.
A clearer side-by-side comparison and demonstration of our system can be found in the accompanying video.
Our system is also significantly faster than other methods, taking less than one second to generate an image.
A detailed timing of each stage is described in \refsec{results:quant}.

\subsection{Quantitative comparison}
\lblsec{results:quant}
We evaluate our method quantitatively by using the test set described in \refsec{implementation:train}. 
We compare with direct video interpolation methods and depth from video methods.
For video interpolation, we interpolate the light field frames using Epicflow~\cite{revaud2015epicflow} and FlowNet~\cite{fischer2015flownet}, independently for each viewpoint.
Note that they only have the light field frames as input, since there is no easy way to incorporate the high speed video frames into their system.
After flows are estimated, the forward warping process is done using the method described in~\cite{baker2011database}.
For depth from video, we compare with depthTransfer~\cite{karsch2014depth}, which generates a depth sequence by finding candidates in the training set.
Their software also enables production of a stereo sequence using the obtained depths.
We thus provide the light field keyframes as training sets, and modify their rendering code to output light fields instead of stereo pairs at each 2D frame.
The average PSNR and SSIM values across all in-between frames and the four corner views are reported in Table~\ref{tab:psnr_system}.
Our method achieves significant improvement over other methods.

\begin{figure}[t!]
  \centering
  \includegraphics[width=.95\linewidth]{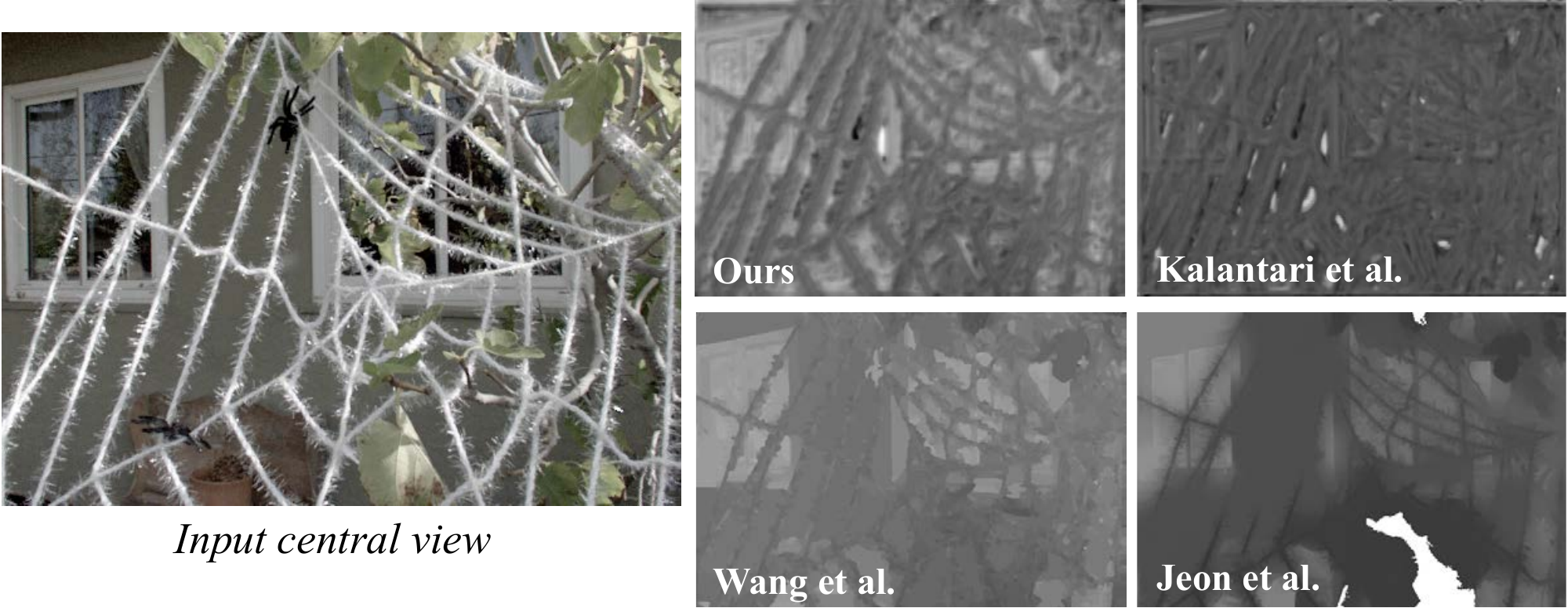}
  \vspace{-.1in}
  \caption{Comparison on depth estimation methods. Note that our depth is more accurate around the occlusion boundaries.}
  \label{fig:depth_compare}
  \vspace{-.1in}
\end{figure}

To evaluate the effectiveness of our system, we also try to replace each component in the system by another method.
The comparison is shown in Table~\ref{tab:psnr_component}.
For the disparity CNN, we replace it with the model in~\cite{kalantari2016learning}, and two other traditional methods~\cite{jeon2015accurate,wang2015occlusion}.
It can be seen that the traditional methods are not optimized for synthesizing views, so their results are much worse;
besides, our method is two to three orders of magnitude faster than these methods.
For~\cite{kalantari2016learning}, although the model is also used for view synthesis, it does not account for temporal flows in the system.
In contrast, our system is trained end-to-end with the flow estimation, thus resulting in better performance.
A visual comparison of generated depth is shown in Fig.~\ref{fig:depth_compare}.
For the temporal flow CNN, we replace it with Epicflow~\cite{revaud2015epicflow} and FlowNet~\cite{fischer2015flownet}.
Although they can in general produce more accurate flows, they are not designed for synthesizing views, so our system still performs slightly better.
Also note that our flow network is two orders of magnitude faster than Epicflow.
For the warp flow CNN, since no easy substitute can be found, we do not have any comparisons.
Finally, for the appearance CNN, we replace it with using only $\widetilde{I^t}$ and a weighted average of all warped images.
Again, these results are not comparable to our original result, which demonstrates the effectiveness of our appearance CNN.

\begin{table}[t!]
\centering
\begin{tabular}{ | c | c | c | c | c | }
\hline
Method & Epicflow & FlowNet &DepthTransfer & Ours \\ \hline
PSNR & 21.35  &  21.70  &  24.88  &  \textbf{32.22} \\ \hline
SSIM & 0.628  &  0.632  &  0.773  & \textbf{0.949} \\ \hline
\end{tabular}
\vspace{.05in}
\caption{Quantitative comparison to 2D video interpolation and depth from video methods.}
\label{tab:psnr_system}
\vspace{-.25in}
\end{table}

\begin{table}[t!]
\centering
\small
\setlength{\tabcolsep}{2pt}
\begin{tabular}{ | c || c | c | c || c | c || c | c | }
\hline
  & \multicolumn{3}{c||}{Disparity} & \multicolumn{2}{c||}{Optical flow}  & \multicolumn{2}{c|}{Appearance} \\ \hline
Method & Kalantari & Jeon & Wang & Epicflow & FlowNet  & $\widetilde{I^t}$ & Avg \\ \hline
PSNR & 29.77 &  26.41  &  25.62  & 31.83  & 31.16 &  30.51 & 31.02 \\ \hline
SSIM & 0.901 & 0.828  &  0.756  & 0.945  &  0.935 &  0.935 & 0.934 \\ \hline
\end{tabular}
\vspace{.05in}
\caption{Component-wise quantitative comparison for our method. 
We replace each component in our system by another method, and let the rest of the system stay the same.
The first row specifies the component of interest; the second row specifies the replacing method.
It can be seen that all methods achieve worse results than our method (shown in Table~\ref{tab:psnr_system}).}
\label{tab:psnr_component}
\vspace{-.25in}
\end{table}

\paragraph{Timing}
Our method takes $0.92$ seconds to generate a $352\times 512$ novel view at a particular timeframe, using an Intel i7 3.4 GHz machine with a GeForce Titan X GPU.
Specifically, it takes $0.66$ seconds to estimate the disparities, $0.2$ seconds to estimate the temporal flows, $0.03$ seconds to evaluate the warp flow CNN and warp the images, and $0.03$ seconds to evaluate the appearance CNN.
Also note that many of these steps (disparity and temporal flow estimations) can be reused when computing different viewpoints, and the parts that require recomputing only take $0.06$ seconds.
In contrast, Epicflow takes about $30$ seconds ($15$ seconds for computing flows and $15$ seconds for warping images using~\cite{baker2011database}), and depthTransfer takes about two minutes.
Although flow estimation by FlowNet is fast (around $0.1$ seconds), the subsequent warping process still takes about $15$ seconds.

\subsection{Qualitative comparison} 
\begin{figure*}[ht!]
  \centering
  \includegraphics[width=.95\linewidth]{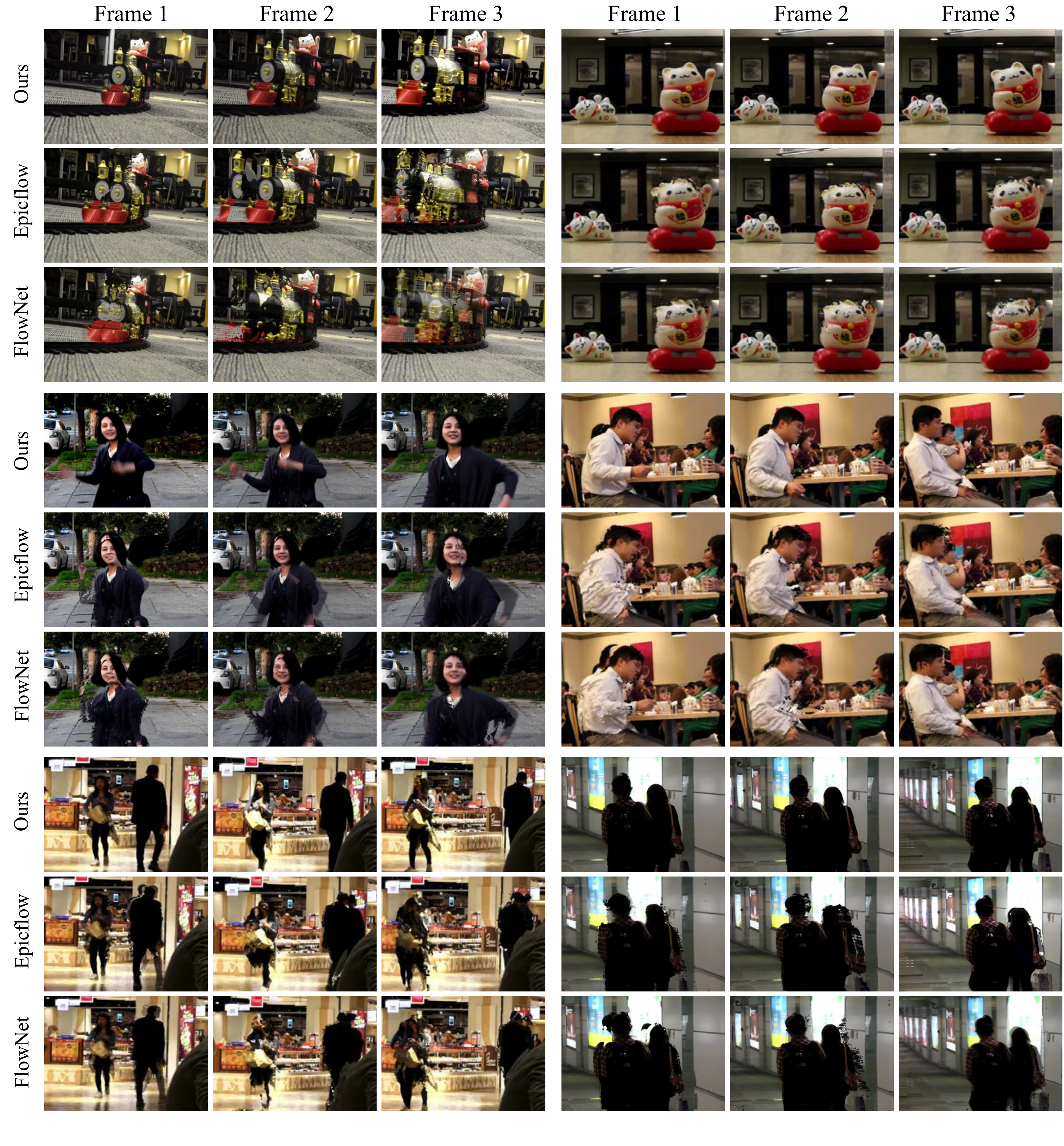}
  \vspace{-.3in}
  \caption{Temporal comparisons against video interpolation methods. 
  For Epicflow and FlowNet, since they lack the 2D video input, they usually produce ghosting or visible artifacts around the moving objects.
  Results are best seen electronically when zoomed in, or in the accompanying video.}
  \label{fig:temporal_compare}
  \vspace{-.1in}
\end{figure*}

\lblsec{results:qual}
We visualize the results taken with our prototype (\reffig{teaser}a) qualitatively in this subsection.
\paragraph{Comparison against video interpolation methods}
For Epicflow and FlowNet, since these methods cannot easily deal with very low fps inputs, they usually produce temporally inconsistent results, as shown in~\reffig{temporal_compare}.
For the first scene, we have a toy train moving toward the camera.
Note that both Epicflow and FlowNet generate ghosting around the train head, especially as the train is closer so it is moving more quickly.
For the lucky cat scene, the body of the cat on the right is swinging.
Although optical flow methods can handle the relatively static areas, they cannot deal with the fast moving head and arm of the cat.
Next, we have a video of a woman dancing. Again, ghosting effects can be seen around the face and the arms of the woman.
Note that in addition to incorrect flows, the ghosting is also partly due to the effect shown in \reffig{occlusion}, so the woman appears to have multiple arms.
For the dining table scene, the man at the front is acting aggressively, shaking his body and waving his arm.
As a result, the other interpolation algorithms fail to produce consistent results.
Finally, for the last two scenes, we capture sequences of walking people, with the camera relatively still in the first scene and moving in the second scene.
Although the people in general are not moving at a high speed, visible artifacts can still be seen around them.

\paragraph{Comparison against depth from video methods}
Comparing with depthTransfer requires more care since it also takes the 2D video as input.
In fact, we observe that it usually just copies or shifts the 2D video frames to produce different angular results.
Therefore, their results seem to be temporally consistent, but they actually lack the parallax which exists in real light field videos.

\begin{figure*}[ht!]
  \centering
  \includegraphics[width=\linewidth]{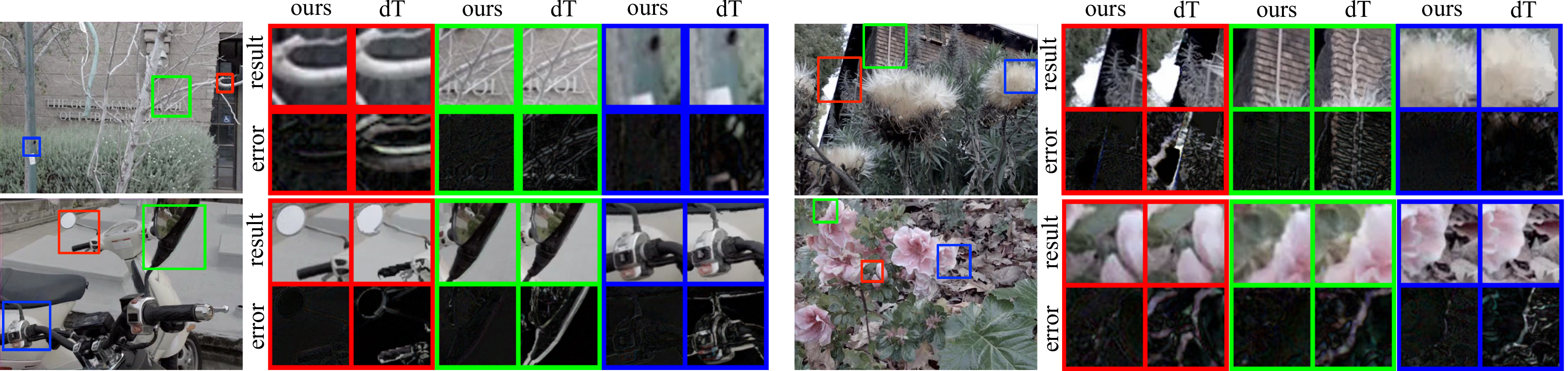}
  \vspace{-.25in}
  \caption{Angular comparisons against depthTransfer (dT) on our test set.
  For each inset, the left column is our result (up) and the error compared to ground truth (bottom), while the right column is the result by depthTransfer and its error.
  The ground truth is not shown here due to its closeness to our result.
  Note that our results have much smaller errors.
  The effect is much easier to see in the accompanying video.}
  \label{fig:quant_compare}
\end{figure*}

To verify the above observation, we first use the test set described in \refsec{implementation:train}, where we have ground truth available.
The results of the upper leftmost angular views produced by both our method and depthTransfer are shown in \reffig{quant_compare}.
For each inset, the left column is our result (above) and the error compared to ground truth (bottom), while the right column is the result by depthTransfer and its error.
For the first scene, we have a tree and a pole in front of a building.
Note that our method realistically reconstructs the structure of the branches.
For depthTransfer, since it usually just shifts the scene, although the result may seem reasonable, it lacks the correct parallax so the difference with the ground truth is still large.
Next, we have a challenging scene of furry flowers.
Note that depthTransfer distorts the building behind and produces artifacts on the flower boundary, while our method can still generate reasonable results.
For the bike scene, again depthTransfer distorts the mirror and fails to generate correct parallax for the steering handle.
Finally, in the last scene our method realistically captures the shape of the flower and the leaves, while depthTransfer has large errors compared to the ground truth.

\begin{figure*}[ht!]
  \centering
  \includegraphics[width=\linewidth]{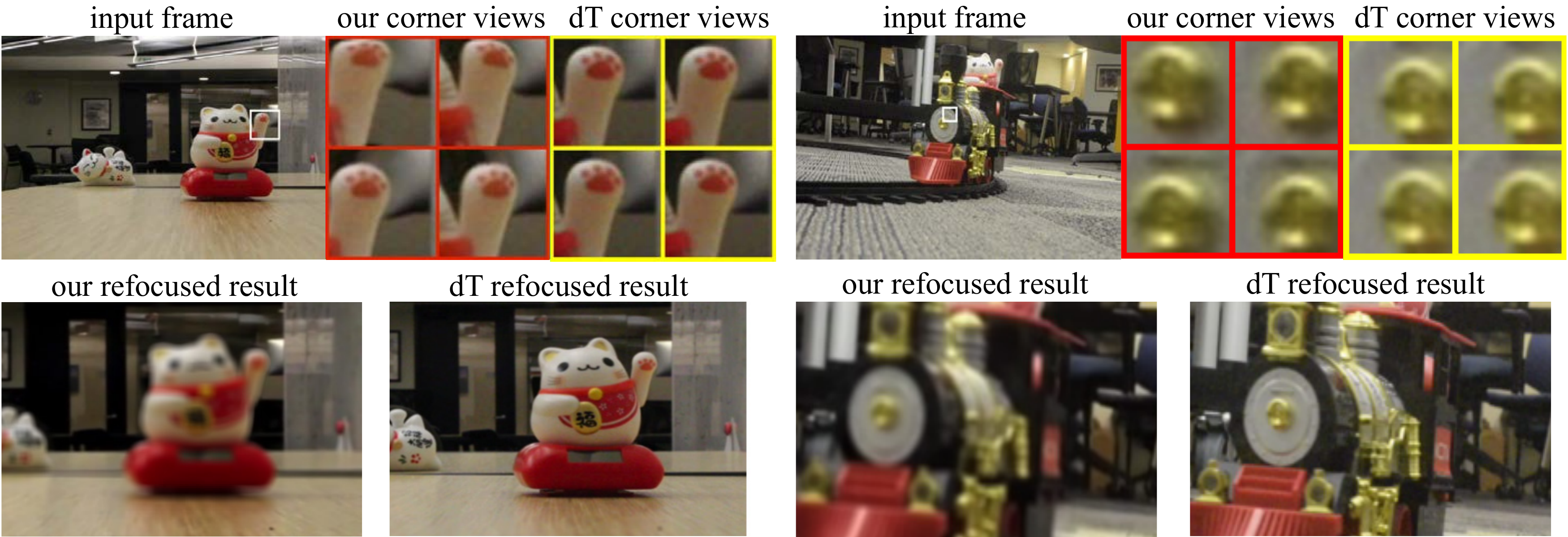}
  \vspace{-.25in}
  \caption{Angular comparisons against depthTransfer (dT) on real-world scenarios.
  In the first row, we show the input frame and the four corner views produced by each method. 
  In the second row, we show the results when refocused to the background.
  It can be seen that our method generates visible parallax between different views, while results by depthTransfer have very little parallax, so their refocused results are sharp everywhere.
  The effect is much easier to see in the accompanying video.}
  \label{fig:angular_compare}
  \vspace{-.05in}
\end{figure*}

Next, we compare with depthTransfer using the scenes captured with our prototype.
Example results are shown in \reffig{angular_compare}.
We show all four corner views at once, so we can see the parallax between different views more easily. 
It can be seen that results by depthTransfer have very little parallax, while our method generates more reasonable results.
This effect is even clearer to see in the refocused images, where results by depthTransfer are actually sharp everywhere.
On the other hand, our refocused images have blurry foreground and sharp background.

\begin{figure}[ht!]
  \centering
    \vspace{-.1in}
  \includegraphics[width=.95\linewidth]{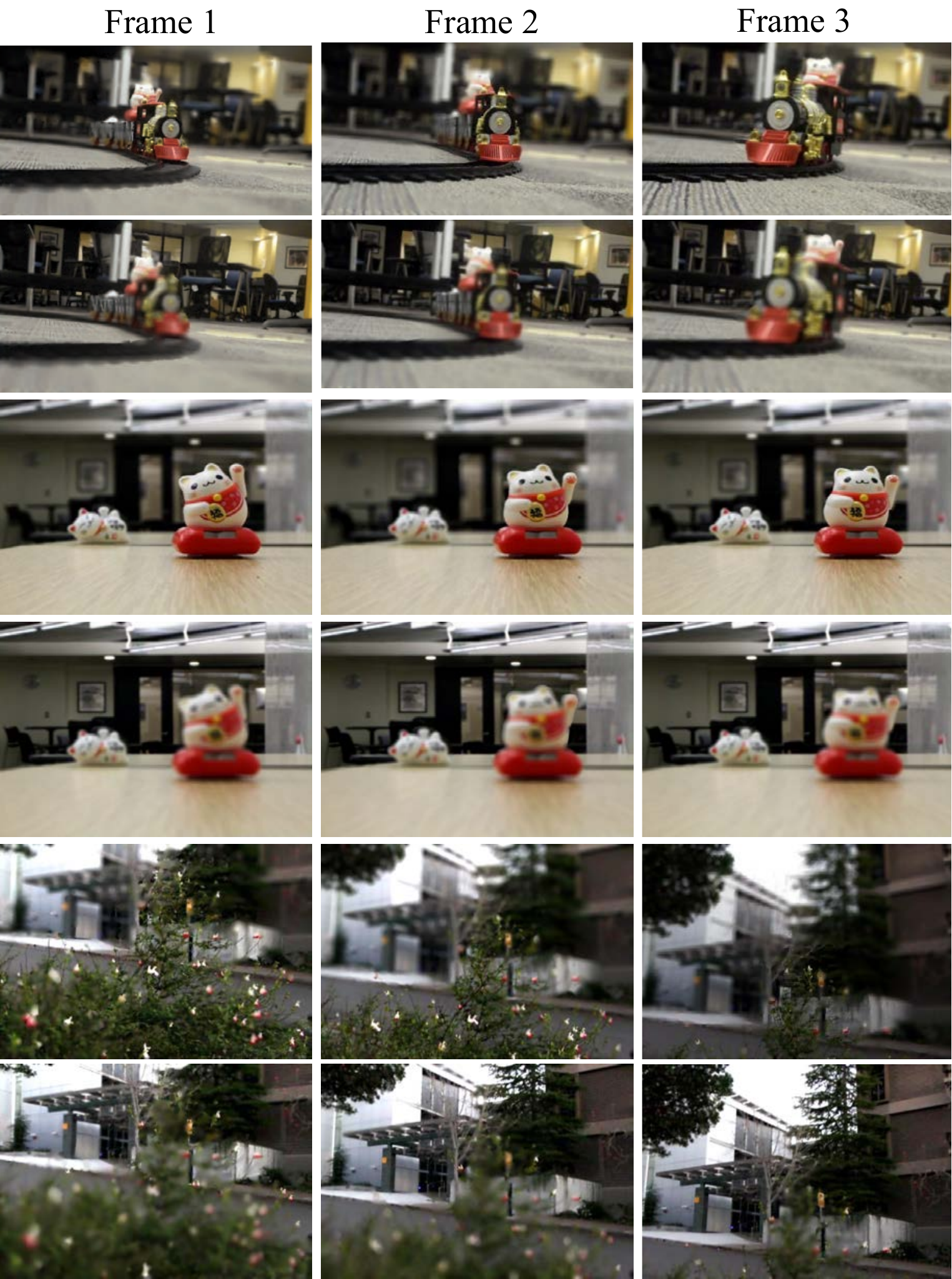}
  \vspace{-.1in}
  \caption{Examples of the video refocusing application. After we get the light fields and disparities at each frame, we can easily change the focus distances throughout the video. 
  For each sequence, the first row shows when it is focused at the front, while the second row shows the video focused at the back.
  Results are best seen electronically when zoomed in.}
  \label{fig:refocus}
  \vspace{-.1in}
\end{figure}

\begin{figure}[ht!]
  \centering
  \includegraphics[width=\linewidth]{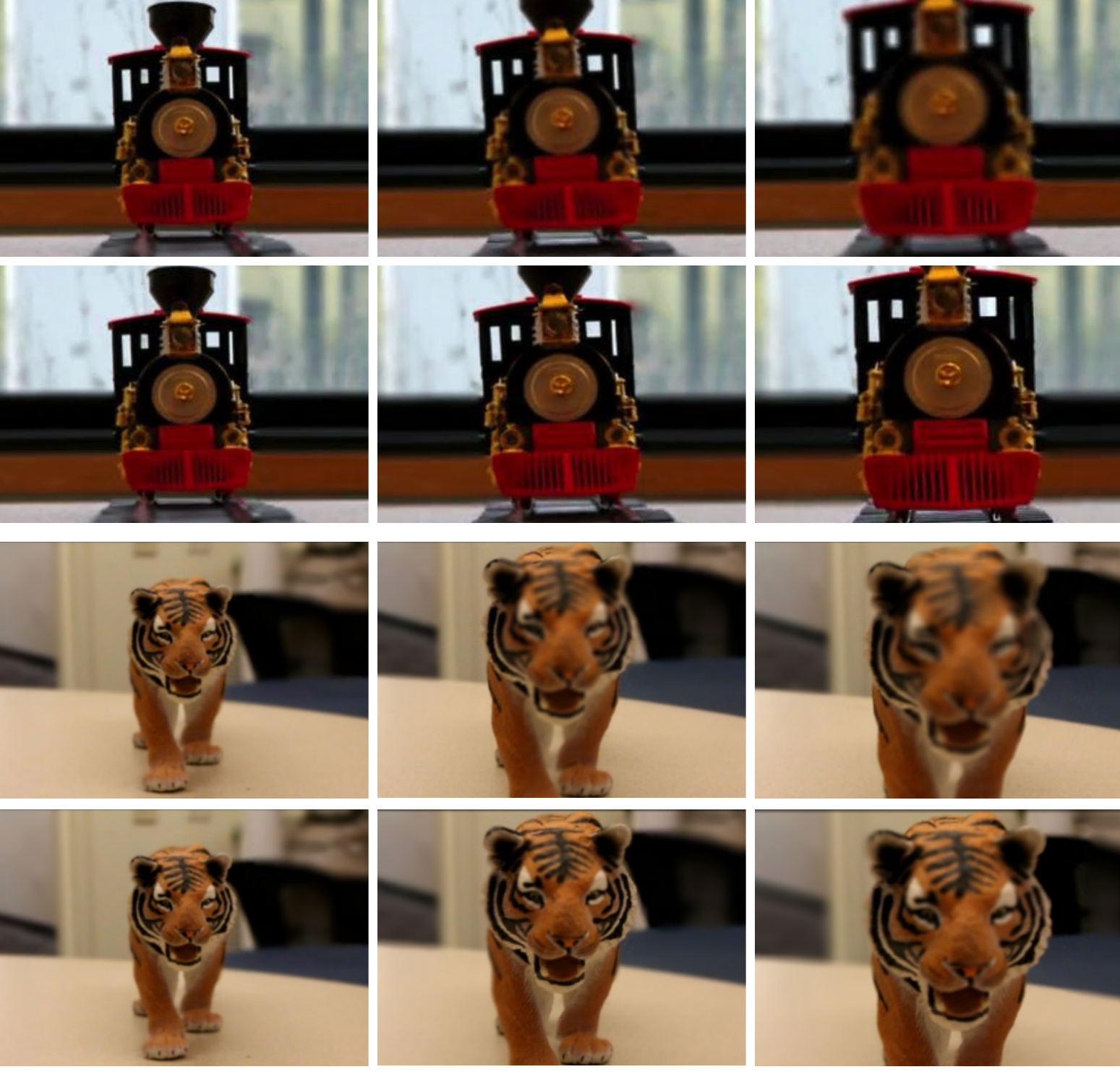}
  \vspace{-.3in}
  \caption{Examples of video focus tracking. In the first row, we fixed the focus plane distance throughout the video. 
  In the second row, we automatically track the point the user clicks at the beginning (the train on the top and the tiger's face in the bottom), and change the focus plane accordingly. }
  \label{fig:focus_track}
  \vspace{-.05in}
\end{figure}

\begin{figure}[ht!]
  \vspace{-.05in}
  \centering
  \includegraphics[width=.95\linewidth]{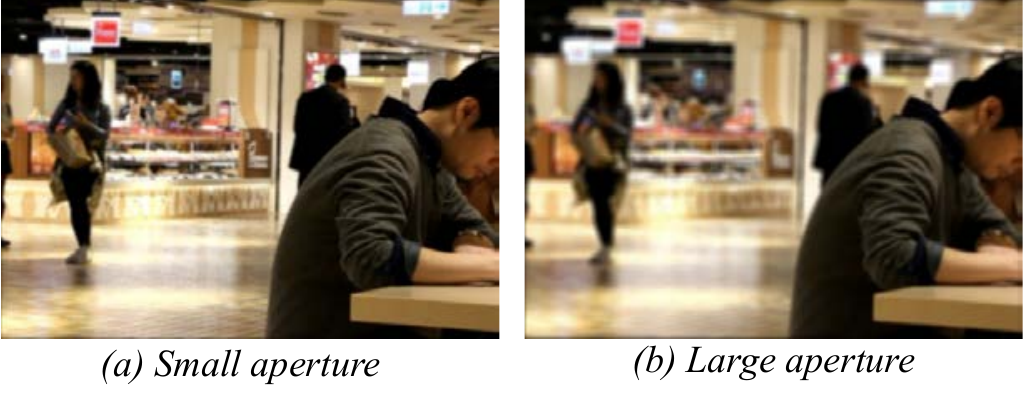}
  \vspace{-.2in}
  \caption{An example of changing aperture. After the video is taken, the depth of field can still easily be changed. In (b) we keep the man in the foreground in focus while blurring the background due to the larger virtual aperture.}
  \label{fig:aperture}
  \vspace{-.1in}
\end{figure}

\subsection{Applications} \lblsec{results:app}
After we obtain the $30$ fps light field video, we demonstrate that we are able to do video refocusing.
This is useful for changing the focus plane back and forth in a video, or fixing the focus on a particular object no matter where it goes.
We also develop an interactive user interface where the user can click on particular points as the video plays, so the point will become in focus.
The usage of the interface is shown in the accompanying video, and the results are shown in \reffig{refocus}.
In an alternative mode, after the user clicks a point at the start of the video, our algorithm automatically tracks that point using KLT and always focuses on it throughout the video; the results are shown in \reffig{focus_track}.
Finally, the user can also change the effective aperture to produce different depths of field at each frame, so that does not need to be determined when shooting the video (\reffig{aperture}).
All these features are only achievable on light field videos, which are captured and rendered for the first time using consumer cameras by our system.

\subsection{Limitations and future work} 
\begin{figure}[ht!]
  \centering
  \includegraphics[width=\linewidth]{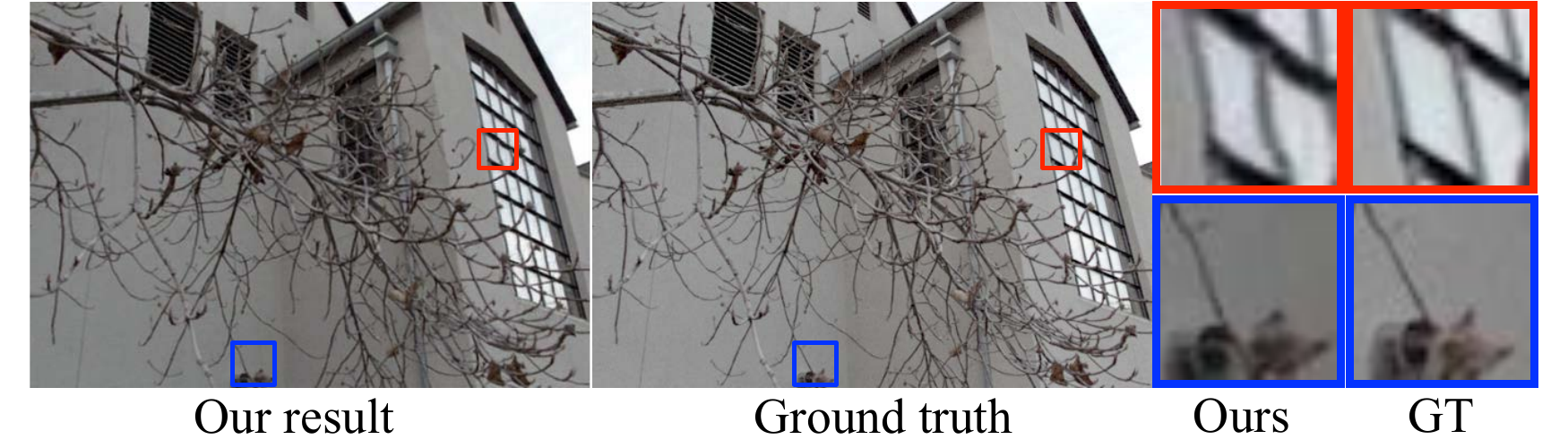}
  \vspace{-.3in}
  \caption{An example on a challenging scene. In the red inset, our method fails to capture the line structure of the window since it was previously occluded by the twig. Similarly, in the blue inset, our method generates elongated leaves compared to the ground truth.}
  \label{fig:failure}
  \vspace{-.1in}
\end{figure}

\lblsec{results:limit}
There are several limitations to our system.
First, our results are still not completely artifact-free. 
We found that artifacts occur mostly around occlusions and are usually caused by improper flow estimation. 
Since flows between images are undefined in occluded regions, images warped by these flows tend to have some errors in the corresponding regions as well. 
Our system improves upon current methods by a large margin thanks to end-to-end training, but still cannot solve this problem entirely. 
An example is shown in Fig.~\ref{fig:failure}.
A benefit of our system is that we decompose the pipeline into different components. In the future, if a better flow method is proposed, we can replace that component with such a new method to improve results.

Second, since the Lytro ILLUM camera has a small baseline, the object we try to shoot cannot be too far away.
However, if the object is too close, the two views from the Lytro camera and the DSLR become too different and will have large occlusions, making the alignment hard.
Therefore, currently our system cannot shoot objects too close or too far away.
This can be mitigated by using a smaller attachable camera other than a DSLR, e.g.\ phone cameras, or having a larger baseline light field camera.
Another more fundamental way to deal with this problem is to integrate this into the learning system as well and train end-to-end, so the network can try to infer the difference between the two cameras.
This may also save us from running NRDC calibration, which is relatively slow.

Third, if motion blur exists in the videos, the flow estimation will often fail, making the disparity transfer fail.
Therefore, currently we ensure the exposure time is low enough that no motion blur occurs.
Usually this is easily achievable as long as the scene is not too dark.
If capturing a dark scene is inevitable, a possible solution is to deblur the images first.
The process can be made easier if we have sharp images from one of the two cameras by using different exposures.
This may also make HDR imaging possible.

Next, we do not utilize the extra resolution in the DSLR now. 
An improvement over our system would be to take that into account and generate a high resolution light field video.
We leave this to future work.

Finally, with the availability of light field videos, we would like to extend previous visual understanding work to videos, such as estimating saliency~\cite{li2014saliency}, materials~\cite{wang20164d}, or matting~\cite{cho2014consistent} for dynamic objects and scenes.

\section{Conclusion}
We propose a novel algorithm to capture a light field video using a hybrid camera system.
Although current light field technologies have many benefits, they are mainly restricted to only still images, due to the high bandwidth usage.
By incorporating an additional 2D camera to capture the temporal information, we propose a pipeline to combine the 2D video and the sparse light field sequence to generate a full light field video.
Our system consists of a spatio-temporal flow CNN and an appearance CNN.
The flow CNN propagates the temporal information from the light field frames to the 2D frames, and warps all images to the target view.
The appearance CNN then takes in these images and generates the final image.
Experimental results demonstrate that our method can capture realistic light field videos, and outperforms current video interpolation and depth from video methods.
This enables many important applications on videos, such as dynamic refocusing and focus tracking.
We believe this is an important new page for light fields, bringing light field imaging techniques to videography.

\appendix
\section{Concatenation of flows} \label{sec:concate}
We show how we warp the five images $I^t$,$I^0$,$I^T$,$L^0$ and $L^T$ to $L^t$ in this section.
In the main text we have already shown how to warp $I^t$ and $I^0$ to get $\widetilde{I^t}$ and $\widetilde{I^0}$ (by (\ref{eq:disparity_backward}) and (\ref{eq:I0_warp})).
We now show how to warp $L^0$.
Let $\textbf{z} \equiv \textbf{y}+f^{0\rightarrow t}(\textbf{y})$.
Then since $I^0(\textbf{x}) = L^0(\textbf{x},\textbf{0}) = L^0(\textbf{x}+\textbf{u}\cdot d^0(\textbf{x}), \textbf{u})$ from (\ref{eq:disparity}), we can rewrite (\ref{eq:I0_warp}) as
\small
\begin{equation}
\begin{aligned}
L^t(\textbf{x}, \textbf{u}) &= I^t(\textbf{y}) = I^0(\textbf{z}) = L^0(\textbf{z}+\textbf{u}\cdot d^0(\textbf{z}), \textbf{u})\\
&= L^0\Big(\textbf{y}+f^{0\rightarrow t}(\textbf{y})+\textbf{u}\cdot d^0\big(\textbf{y}+f^{0\rightarrow t}(\textbf{y})\big), \textbf{u}\Big)\\
\equiv \widetilde{L^0} &= L^0\Big(\textbf{x}-\textbf{u}\cdot d_{\textbf{u}}^t(\textbf{x})+f^{0\rightarrow t}(\textbf{x}-\textbf{u}\cdot d_{\textbf{u}}^t(\textbf{x}))+ \\
& \quad\quad \textbf{u}\cdot d^0\big(\textbf{x}-\textbf{u}\cdot d_{\textbf{u}}^t(\textbf{x})+f^{0\rightarrow t}(\textbf{x}-\textbf{u}\cdot d_{\textbf{u}}^t(\textbf{x}))\big), \textbf{u}\Big)\\
\end{aligned}
\end{equation}
\normalsize

Similarly, we can write the warp flows for $I^T$ and $L^T$ as
\small
\begin{equation}
\widetilde{I^T}(\textbf{x}) = I^T\big(\textbf{x}-\textbf{u}\cdot d_{\textbf{u}}^t(\textbf{x}) +f^{T\rightarrow t}(\textbf{x}-\textbf{u}\cdot d_{\textbf{u}}^t(\textbf{x}))\big)
\end{equation}
\begin{equation}
\begin{aligned}
\widetilde{L^T}(\textbf{x}, \textbf{u}) = &L^T\Big(\textbf{x}-\textbf{u}\cdot d_{\textbf{u}}^t(\textbf{x})+f^{T\rightarrow t}(\textbf{x}-\textbf{u}\cdot d_{\textbf{u}}^t(\textbf{x}))+ \\
&\textbf{u}\cdot d^T\big(\textbf{x}-\textbf{u}\cdot d_{\textbf{u}}^t(\textbf{x})+f^{T\rightarrow t}(\textbf{x}-\textbf{u}\cdot d_{\textbf{u}}^t(\textbf{x}))\big), \textbf{u}\Big)\\
\end{aligned}
\end{equation}
\normalsize

\section{Training details} \lblsec{training}
For the end-to-end training, ideally a training example should contain all the images between two key frames and try to output the entire light field sequence.
That is, we should have the two light fields $(L^0, L^T)$ (which include $I^0$ and $I^T$) and the 2D frames $\{I^1, ..., I^{t-1}\}$ as inputs, and try to output $\{L^1, ..., L^{t-1}\}$.
However, as it is not possible to fit the entire light field sequence into memory during training, each training example only samples some views in the light fields and one frame in the 2D sequence.
In particular, instead of using all angular views in the two key frames $0$ and $T$, each time we only randomly sample 4 views $\textbf{u}_1,\textbf{u}_2,\textbf{u}_3,\textbf{u}_4$ in addition to the central view $\textbf{u}_0$.
These five views are then used in the disparity estimation network to generate the disparity at the key frames.
Similarly, instead of using the central views of all in-between frames, for each training example we only select one frame $t$ as input.
The flows between the key frames and frame $t$ are then estimated using the temporal flow network.
This can be seen as angular and temporal ``crop,'' just as we would randomly crop in the spatial domain during training.
For output, we also randomly sample one angular view $\textbf{u}$ at frame $t$.
The warp flow network then takes view $\textbf{u}$ and $\textbf{u}_0$ (the central view) at the two key frames, as well as the 2D frame $I^t$ and warps them.
Finally the color estimation network takes these warped images and generates the final output.
More clearly, the inputs for one training example are: six views ($\textbf{u}_0, ..., \textbf{u}_4$ and $\textbf{u}$) in the two key frame light fields $L^0$ and $L^T$, and one 2D frame $I^t$ ($I^0$ and $I^T$ are just view $\textbf{u}_0$ from $L^0$ and $L^T$). 
The output $L^t(\textbf{x},\textbf{u})$ is one view at frame $t$.
The network then tries to minimize the image reconstruction loss (measured by Euclidean distance) between the generated output and $L^t(\textbf{x},\textbf{u})$.

\section*{Acknowledgements}
We would like to gratefully thank Manmohan Chandraker and Ren Ng for valuable discussions.
This work was funded in part by ONR grant N00014152013, NSF grants 1451830, 1617234 and 1539099, a Google research award, a Facebook fellowship, and the UC San Diego Center for Visual Computing.

\bibliography{main}


\begin{thebibliography}{00}

%%% the Web address.

\ifx \showCODEN    \undefined \def \showCODEN     #1{\unskip}     \fi
\ifx \showDOI      \undefined \def \showDOI       #1{{\tt DOI:}\penalty0{#1}\ }
  \fi
\ifx \showISBNx    \undefined \def \showISBNx     #1{\unskip}     \fi
\ifx \showISBNxiii \undefined \def \showISBNxiii  #1{\unskip}     \fi
\ifx \showISSN     \undefined \def \showISSN      #1{\unskip}     \fi
\ifx \showLCCN     \undefined \def \showLCCN      #1{\unskip}     \fi
\ifx \shownote     \undefined \def \shownote      #1{#1}          \fi
\ifx \showarticletitle \undefined \def \showarticletitle #1{#1}   \fi
\ifx \showURL      \undefined \def \showURL       {\relax}        \fi
% invisible to TeX
\providecommand\bibfield[2]{#2}
\providecommand\bibinfo[2]{#2}
\providecommand\natexlab[1]{#1}
\providecommand\showeprint[2][]{arXiv:#2}

\bibitem[\protect\citeauthoryear{Baker, Scharstein, Lewis, Roth, Black, and
  Szeliski}{Baker et~al\mbox{.}}{2011}]%
        {baker2011database}
\bibfield{author}{\bibinfo{person}{Simon Baker}, \bibinfo{person}{Daniel
  Scharstein}, \bibinfo{person}{JP Lewis}, \bibinfo{person}{Stefan Roth},
  \bibinfo{person}{Michael~J Black}, {and} \bibinfo{person}{Richard Szeliski}.}
  \bibinfo{year}{2011}\natexlab{}.
\newblock \showarticletitle{A database and evaluation methodology for optical
  flow}.
\newblock \bibinfo{journal}{{\em International Journal of Computer Vision
  (IJCV)\/}} \bibinfo{volume}{92}, \bibinfo{number}{1} (\bibinfo{year}{2011}),
  \bibinfo{pages}{1--31}.
\newblock


\bibitem[\protect\citeauthoryear{Ben-Ezra and Nayar}{Ben-Ezra and
  Nayar}{2003}]%
        {ben2003motion}
\bibfield{author}{\bibinfo{person}{Moshe Ben-Ezra} {and}
  \bibinfo{person}{Shree~K Nayar}.} \bibinfo{year}{2003}\natexlab{}.
\newblock \showarticletitle{Motion deblurring using hybrid imaging}. In
  \bibinfo{booktitle}{{\em IEEE Conference on Computer Vision and Pattern
  Recognition (CVPR)}}, Vol.~\bibinfo{volume}{1}. \bibinfo{pages}{I--657}.
\newblock


\bibitem[\protect\citeauthoryear{Bhat, Zitnick, Snavely, Agarwala, Agrawala,
  Cohen, Curless, and Kang}{Bhat et~al\mbox{.}}{2007}]%
        {bhat2007using}
\bibfield{author}{\bibinfo{person}{Pravin Bhat}, \bibinfo{person}{C~Lawrence
  Zitnick}, \bibinfo{person}{Noah Snavely}, \bibinfo{person}{Aseem Agarwala},
  \bibinfo{person}{Maneesh Agrawala}, \bibinfo{person}{Michael Cohen},
  \bibinfo{person}{Brian Curless}, {and} \bibinfo{person}{Sing~Bing Kang}.}
  \bibinfo{year}{2007}\natexlab{}.
\newblock \showarticletitle{Using photographs to enhance videos of a static
  scene}. In \bibinfo{booktitle}{{\em Eurographics Symposium on Rendering
  (EGSR)}}. \bibinfo{pages}{327--338}.
\newblock


\bibitem[\protect\citeauthoryear{Bishop, Zanetti, and Favaro}{Bishop
  et~al\mbox{.}}{2009}]%
        {bishop2009light}
\bibfield{author}{\bibinfo{person}{Tom~E Bishop}, \bibinfo{person}{Sara
  Zanetti}, {and} \bibinfo{person}{Paolo Favaro}.}
  \bibinfo{year}{2009}\natexlab{}.
\newblock \showarticletitle{Light field superresolution}. In
  \bibinfo{booktitle}{{\em IEEE International Conference on Computational
  Photography (ICCP)}}. \bibinfo{pages}{1--9}.
\newblock


\bibitem[\protect\citeauthoryear{Boominathan, Mitra, and
  Veeraraghavan}{Boominathan et~al\mbox{.}}{2014}]%
        {boominathan2014improving}
\bibfield{author}{\bibinfo{person}{Vivek Boominathan}, \bibinfo{person}{Kaushik
  Mitra}, {and} \bibinfo{person}{Ashok Veeraraghavan}.}
  \bibinfo{year}{2014}\natexlab{}.
\newblock \showarticletitle{Improving resolution and depth-of-field of light
  field cameras using a hybrid imaging system}. In \bibinfo{booktitle}{{\em
  IEEE International Conference on Computational Photography (ICCP)}}.
  \bibinfo{pages}{1--10}.
\newblock


\bibitem[\protect\citeauthoryear{Cao, Tong, Dai, and Lin}{Cao
  et~al\mbox{.}}{2011}]%
        {cao2011high}
\bibfield{author}{\bibinfo{person}{Xun Cao}, \bibinfo{person}{Xin Tong},
  \bibinfo{person}{Qionghai Dai}, {and} \bibinfo{person}{Stephen Lin}.}
  \bibinfo{year}{2011}\natexlab{}.
\newblock \showarticletitle{High resolution multispectral video capture with a
  hybrid camera system}. In \bibinfo{booktitle}{{\em IEEE Conference on
  Computer Vision and Pattern Recognition (CVPR)}}. \bibinfo{pages}{297--304}.
\newblock


\bibitem[\protect\citeauthoryear{Chaurasia, Duchene, Sorkine-Hornung, and
  Drettakis}{Chaurasia et~al\mbox{.}}{2013}]%
        {chaurasia2013depth}
\bibfield{author}{\bibinfo{person}{Gaurav Chaurasia}, \bibinfo{person}{Sylvain
  Duchene}, \bibinfo{person}{Olga Sorkine-Hornung}, {and}
  \bibinfo{person}{George Drettakis}.} \bibinfo{year}{2013}\natexlab{}.
\newblock \showarticletitle{Depth Synthesis and Local Warps for Plausible
  Image-based Navigation}.
\newblock \bibinfo{journal}{{\em ACM Transactions on Graphics (TOG)\/}}
  \bibinfo{volume}{32}, \bibinfo{number}{3} (\bibinfo{year}{2013}),
  \bibinfo{pages}{30:1--30:12}.
\newblock


\bibitem[\protect\citeauthoryear{Cho, Kim, and Tai}{Cho et~al\mbox{.}}{2014}]%
        {cho2014consistent}
\bibfield{author}{\bibinfo{person}{Donghyeon Cho}, \bibinfo{person}{Sunyeong
  Kim}, {and} \bibinfo{person}{Yu-Wing Tai}.} \bibinfo{year}{2014}\natexlab{}.
\newblock \showarticletitle{Consistent matting for light field images}. In
  \bibinfo{booktitle}{{\em European Conference on Computer Vision (ECCV)}}.
  \bibinfo{pages}{90--104}.
\newblock


\bibitem[\protect\citeauthoryear{Cho, Lee, Kim, and Tai}{Cho
  et~al\mbox{.}}{2013}]%
        {cho2013modeling}
\bibfield{author}{\bibinfo{person}{Donghyeon Cho}, \bibinfo{person}{Minhaeng
  Lee}, \bibinfo{person}{Sunyeong Kim}, {and} \bibinfo{person}{Yu-Wing Tai}.}
  \bibinfo{year}{2013}\natexlab{}.
\newblock \showarticletitle{Modeling the calibration pipeline of the lytro
  camera for high quality light-field image reconstruction}. In
  \bibinfo{booktitle}{{\em IEEE International Conference on Computer Vision
  (ICCV)}}. \bibinfo{pages}{3280--3287}.
\newblock


\bibitem[\protect\citeauthoryear{Dosovitskiy, Fischery, Ilg, Haz{\i}rba{\c{s}},
  Golkov, van~der Smagt, Cremers, and Brox}{Dosovitskiy et~al\mbox{.}}{2015}]%
        {fischer2015flownet}
\bibfield{author}{\bibinfo{person}{Alexey Dosovitskiy},
  \bibinfo{person}{Philipp Fischery}, \bibinfo{person}{Eddy Ilg},
  \bibinfo{person}{Caner Haz{\i}rba{\c{s}}}, \bibinfo{person}{Vladimir Golkov},
  \bibinfo{person}{Patrick van~der Smagt}, \bibinfo{person}{Daniel Cremers},
  {and} \bibinfo{person}{Thomas Brox}.} \bibinfo{year}{2015}\natexlab{}.
\newblock \showarticletitle{Flownet: Learning optical flow with convolutional
  networks}. In \bibinfo{booktitle}{{\em IEEE International Conference on
  Computer Vision (ICCV)}}. \bibinfo{pages}{2758--2766}.
\newblock


\bibitem[\protect\citeauthoryear{Eisemann, De~Decker, Magnor, Bekaert,
  De~Aguiar, Ahmed, Theobalt, and Sellent}{Eisemann et~al\mbox{.}}{2008}]%
        {eisemann2008floating}
\bibfield{author}{\bibinfo{person}{Martin Eisemann}, \bibinfo{person}{Bert
  De~Decker}, \bibinfo{person}{Marcus Magnor}, \bibinfo{person}{Philippe
  Bekaert}, \bibinfo{person}{Edilson De~Aguiar}, \bibinfo{person}{Naveed
  Ahmed}, \bibinfo{person}{Christian Theobalt}, {and} \bibinfo{person}{Anita
  Sellent}.} \bibinfo{year}{2008}\natexlab{}.
\newblock \showarticletitle{Floating Textures}.
\newblock \bibinfo{journal}{{\em Computer Graphics Forum\/}}
  \bibinfo{volume}{27}, \bibinfo{number}{2} (\bibinfo{year}{2008}),
  \bibinfo{pages}{409--418}.
\newblock


\bibitem[\protect\citeauthoryear{Flynn, Neulander, Philbin, and Snavely}{Flynn
  et~al\mbox{.}}{2016}]%
        {flynn2015deepstereo}
\bibfield{author}{\bibinfo{person}{John Flynn}, \bibinfo{person}{Ivan
  Neulander}, \bibinfo{person}{James Philbin}, {and} \bibinfo{person}{Noah
  Snavely}.} \bibinfo{year}{2016}\natexlab{}.
\newblock \showarticletitle{DeepStereo: Learning to Predict New Views from the
  World's Imagery}. In \bibinfo{booktitle}{{\em IEEE Conference on Computer
  Vision and Pattern Recognition (CVPR)}}. \bibinfo{pages}{5515--5524}.
\newblock


\bibitem[\protect\citeauthoryear{Goesele, Ackermann, Fuhrmann, Haubold,
  Klowsky, Steedly, and Szeliski}{Goesele et~al\mbox{.}}{2010}]%
        {goesele2010ambient}
\bibfield{author}{\bibinfo{person}{Michael Goesele}, \bibinfo{person}{Jens
  Ackermann}, \bibinfo{person}{Simon Fuhrmann}, \bibinfo{person}{Carsten
  Haubold}, \bibinfo{person}{Ronny Klowsky}, \bibinfo{person}{Drew Steedly},
  {and} \bibinfo{person}{Richard Szeliski}.} \bibinfo{year}{2010}\natexlab{}.
\newblock \showarticletitle{Ambient point clouds for view interpolation}.
\newblock \bibinfo{journal}{{\em ACM Transactions on Graphics (TOG)\/}}
  \bibinfo{volume}{29}, \bibinfo{number}{4} (\bibinfo{year}{2010}),
  \bibinfo{pages}{95}.
\newblock


\bibitem[\protect\citeauthoryear{HaCohen, Shechtman, Goldman, and
  Lischinski}{HaCohen et~al\mbox{.}}{2011}]%
        {hacohen2011non}
\bibfield{author}{\bibinfo{person}{Yoav HaCohen}, \bibinfo{person}{Eli
  Shechtman}, \bibinfo{person}{Dan~B Goldman}, {and} \bibinfo{person}{Dani
  Lischinski}.} \bibinfo{year}{2011}\natexlab{}.
\newblock \showarticletitle{Non-rigid dense correspondence with applications
  for image enhancement}.
\newblock \bibinfo{journal}{{\em ACM Transactions on Graphics (TOG)\/}}
  \bibinfo{volume}{30}, \bibinfo{number}{4} (\bibinfo{year}{2011}),
  \bibinfo{pages}{70}.
\newblock


\bibitem[\protect\citeauthoryear{He, Zhang, Ren, and Sun}{He
  et~al\mbox{.}}{2015}]%
        {he2015delving}
\bibfield{author}{\bibinfo{person}{Kaiming He}, \bibinfo{person}{Xiangyu
  Zhang}, \bibinfo{person}{Shaoqing Ren}, {and} \bibinfo{person}{Jian Sun}.}
  \bibinfo{year}{2015}\natexlab{}.
\newblock \showarticletitle{Delving deep into rectifiers: Surpassing
  human-level performance on imagenet classification}. In
  \bibinfo{booktitle}{{\em IEEE International Conference on Computer Vision
  (ICCV)}}. \bibinfo{pages}{1026--1034}.
\newblock


\bibitem[\protect\citeauthoryear{Hinton and Salakhutdinov}{Hinton and
  Salakhutdinov}{2006}]%
        {hinton2006reducing}
\bibfield{author}{\bibinfo{person}{Geoffrey~E Hinton} {and}
  \bibinfo{person}{Ruslan~R Salakhutdinov}.} \bibinfo{year}{2006}\natexlab{}.
\newblock \showarticletitle{Reducing the dimensionality of data with neural
  networks}.
\newblock \bibinfo{journal}{{\em Science\/}} \bibinfo{volume}{313},
  \bibinfo{number}{5786} (\bibinfo{year}{2006}), \bibinfo{pages}{504--507}.
\newblock


\bibitem[\protect\citeauthoryear{Jeon, Park, Choe, Park, Bok, Tai, and
  Kweon}{Jeon et~al\mbox{.}}{2015}]%
        {jeon2015accurate}
\bibfield{author}{\bibinfo{person}{Hae-Gon Jeon}, \bibinfo{person}{Jaesik
  Park}, \bibinfo{person}{Gyeongmin Choe}, \bibinfo{person}{Jinsun Park},
  \bibinfo{person}{Yunsu Bok}, \bibinfo{person}{Yu-Wing Tai}, {and}
  \bibinfo{person}{In~So Kweon}.} \bibinfo{year}{2015}\natexlab{}.
\newblock \showarticletitle{Accurate depth map estimation from a lenslet light
  field camera}. In \bibinfo{booktitle}{{\em IEEE Conference on Computer Vision
  and Pattern Recognition (CVPR)}}. \bibinfo{pages}{1547--1555}.
\newblock


\bibitem[\protect\citeauthoryear{Kalantari, Wang, and Ramamoorthi}{Kalantari
  et~al\mbox{.}}{2016}]%
        {kalantari2016learning}
\bibfield{author}{\bibinfo{person}{Nima~Khademi Kalantari},
  \bibinfo{person}{Ting-Chun Wang}, {and} \bibinfo{person}{Ravi Ramamoorthi}.}
  \bibinfo{year}{2016}\natexlab{}.
\newblock \showarticletitle{Learning-based view synthesis for light field
  cameras}.
\newblock \bibinfo{journal}{{\em ACM Transactions on Graphics (TOG)\/}}
  \bibinfo{volume}{35}, \bibinfo{number}{6} (\bibinfo{year}{2016}),
  \bibinfo{pages}{193}.
\newblock


\bibitem[\protect\citeauthoryear{Karsch, Liu, and Kang}{Karsch
  et~al\mbox{.}}{2014}]%
        {karsch2014depth}
\bibfield{author}{\bibinfo{person}{Kevin Karsch}, \bibinfo{person}{Ce Liu},
  {and} \bibinfo{person}{Sing~Bing Kang}.} \bibinfo{year}{2014}\natexlab{}.
\newblock \showarticletitle{DepthTransfer: Depth extraction from video using
  non-parametric sampling}.
\newblock \bibinfo{journal}{{\em IEEE Transactions on Pattern Analysis and
  Machine Intelligence (TPAMI)\/}} \bibinfo{volume}{36}, \bibinfo{number}{11}
  (\bibinfo{year}{2014}), \bibinfo{pages}{2144--2158}.
\newblock


\bibitem[\protect\citeauthoryear{Kawakami, Matsushita, Wright, Ben-Ezra, Tai,
  and Ikeuchi}{Kawakami et~al\mbox{.}}{2011}]%
        {kawakami2011high}
\bibfield{author}{\bibinfo{person}{Rei Kawakami}, \bibinfo{person}{Yasuyuki
  Matsushita}, \bibinfo{person}{John Wright}, \bibinfo{person}{Moshe Ben-Ezra},
  \bibinfo{person}{Yu-Wing Tai}, {and} \bibinfo{person}{Katsushi Ikeuchi}.}
  \bibinfo{year}{2011}\natexlab{}.
\newblock \showarticletitle{High-resolution hyperspectral imaging via matrix
  factorization}. In \bibinfo{booktitle}{{\em IEEE Conference on Computer
  Vision and Pattern Recognition (CVPR)}}. \bibinfo{pages}{2329--2336}.
\newblock


\bibitem[\protect\citeauthoryear{Kingma and Ba}{Kingma and Ba}{2014}]%
        {kingma2014adam}
\bibfield{author}{\bibinfo{person}{Diederik Kingma} {and}
  \bibinfo{person}{Jimmy Ba}.} \bibinfo{year}{2014}\natexlab{}.
\newblock \showarticletitle{Adam: A method for stochastic optimization}.
\newblock \bibinfo{journal}{{\em arXiv preprint arXiv:1412.6980\/}}
  (\bibinfo{year}{2014}).
\newblock


\bibitem[\protect\citeauthoryear{Konrad, Wang, and Ishwar}{Konrad
  et~al\mbox{.}}{2012}]%
        {konrad20122d}
\bibfield{author}{\bibinfo{person}{Janusz Konrad}, \bibinfo{person}{Meng Wang},
  {and} \bibinfo{person}{Prakash Ishwar}.} \bibinfo{year}{2012}\natexlab{}.
\newblock \showarticletitle{{2D}-to-{3D} image conversion by learning depth
  from examples}. In \bibinfo{booktitle}{{\em IEEE Conference on Computer
  Vision and Pattern Recognition (CVPR) Workshop}}. \bibinfo{pages}{16--22}.
\newblock


\bibitem[\protect\citeauthoryear{Konrad, Wang, Ishwar, Wu, and
  Mukherjee}{Konrad et~al\mbox{.}}{2013}]%
        {konrad2013learning}
\bibfield{author}{\bibinfo{person}{Janusz Konrad}, \bibinfo{person}{Meng Wang},
  \bibinfo{person}{Prakash Ishwar}, \bibinfo{person}{Chen Wu}, {and}
  \bibinfo{person}{Debargha Mukherjee}.} \bibinfo{year}{2013}\natexlab{}.
\newblock \showarticletitle{Learning-based, automatic {2D}-to-{3D} image and
  video conversion}.
\newblock \bibinfo{journal}{{\em IEEE Transactions on Image Processing
  (TIP)\/}} \bibinfo{volume}{22}, \bibinfo{number}{9} (\bibinfo{year}{2013}),
  \bibinfo{pages}{3485--3496}.
\newblock


\bibitem[\protect\citeauthoryear{LeCun, Bottou, Bengio, and Haffner}{LeCun
  et~al\mbox{.}}{1998}]%
        {lecun1998gradient}
\bibfield{author}{\bibinfo{person}{Yann LeCun}, \bibinfo{person}{L{\'e}on
  Bottou}, \bibinfo{person}{Yoshua Bengio}, {and} \bibinfo{person}{Patrick
  Haffner}.} \bibinfo{year}{1998}\natexlab{}.
\newblock \showarticletitle{Gradient-based learning applied to document
  recognition}.
\newblock \bibinfo{journal}{{\it Proc. IEEE}} \bibinfo{volume}{86},
  \bibinfo{number}{11} (\bibinfo{year}{1998}), \bibinfo{pages}{2278--2324}.
\newblock


\bibitem[\protect\citeauthoryear{Levin and Durand}{Levin and Durand}{2010}]%
        {levin2010linear}
\bibfield{author}{\bibinfo{person}{Anat Levin} {and} \bibinfo{person}{Fredo
  Durand}.} \bibinfo{year}{2010}\natexlab{}.
\newblock \showarticletitle{Linear view synthesis using a dimensionality gap
  light field prior}. In \bibinfo{booktitle}{{\em IEEE Conference on Computer
  Vision and Pattern Recognition (CVPR)}}. \bibinfo{pages}{1831--1838}.
\newblock


\bibitem[\protect\citeauthoryear{Li, Ye, Ji, Ling, and Yu}{Li
  et~al\mbox{.}}{2014}]%
        {li2014saliency}
\bibfield{author}{\bibinfo{person}{Nianyi Li}, \bibinfo{person}{Jinwei Ye},
  \bibinfo{person}{Yu Ji}, \bibinfo{person}{Haibin Ling}, {and}
  \bibinfo{person}{Jingyi Yu}.} \bibinfo{year}{2014}\natexlab{}.
\newblock \showarticletitle{Saliency detection on light field}. In
  \bibinfo{booktitle}{{\em IEEE Conference on Computer Vision and Pattern
  Recognition (CVPR)}}. \bibinfo{pages}{2806--2813}.
\newblock


\bibitem[\protect\citeauthoryear{Liao, Lima, Nehab, Hoppe, Sander, and Yu}{Liao
  et~al\mbox{.}}{2014}]%
        {liao2014automating}
\bibfield{author}{\bibinfo{person}{Jing Liao}, \bibinfo{person}{Rodolfo~S
  Lima}, \bibinfo{person}{Diego Nehab}, \bibinfo{person}{Hugues Hoppe},
  \bibinfo{person}{Pedro~V Sander}, {and} \bibinfo{person}{Jinhui Yu}.}
  \bibinfo{year}{2014}\natexlab{}.
\newblock \showarticletitle{Automating image morphing using structural
  similarity on a halfway domain}.
\newblock \bibinfo{journal}{{\em ACM Transactions on Graphics (TOG)\/}}
  \bibinfo{volume}{33}, \bibinfo{number}{5} (\bibinfo{year}{2014}),
  \bibinfo{pages}{168}.
\newblock


\bibitem[\protect\citeauthoryear{{Lytro Cinema}}{{Lytro Cinema}}{2017}]%
        {Lytro}
\bibfield{author}{\bibinfo{person}{{Lytro Cinema}}.}
  \bibinfo{year}{2017}\natexlab{}.
\newblock \bibinfo{title}{The ultimate creative tool for cinema and broadcast}.
\newblock \bibinfo{howpublished}{\url{https://www.lytro.com/cinema}}.
  (\bibinfo{year}{2017}).
\newblock


\bibitem[\protect\citeauthoryear{Mahajan, Huang, Matusik, Ramamoorthi, and
  Belhumeur}{Mahajan et~al\mbox{.}}{2009}]%
        {mahajan2009moving}
\bibfield{author}{\bibinfo{person}{Dhruv Mahajan}, \bibinfo{person}{Fu-Chung
  Huang}, \bibinfo{person}{Wojciech Matusik}, \bibinfo{person}{Ravi
  Ramamoorthi}, {and} \bibinfo{person}{Peter Belhumeur}.}
  \bibinfo{year}{2009}\natexlab{}.
\newblock \showarticletitle{Moving gradients: a path-based method for plausible
  image interpolation}.
\newblock \bibinfo{journal}{{\em ACM Transactions on Graphics (TOG)\/}}
  \bibinfo{volume}{28}, \bibinfo{number}{3} (\bibinfo{year}{2009}),
  \bibinfo{pages}{42}.
\newblock


\bibitem[\protect\citeauthoryear{Marwah, Wetzstein, Bando, and Raskar}{Marwah
  et~al\mbox{.}}{2013}]%
        {marwah2013compressive}
\bibfield{author}{\bibinfo{person}{Kshitij Marwah}, \bibinfo{person}{Gordon
  Wetzstein}, \bibinfo{person}{Yosuke Bando}, {and} \bibinfo{person}{Ramesh
  Raskar}.} \bibinfo{year}{2013}\natexlab{}.
\newblock \showarticletitle{Compressive Light Field Photography Using
  Overcomplete Dictionaries and Optimized Projections}.
\newblock \bibinfo{journal}{{\em ACM Transactions on Graphics (TOG)\/}}
  \bibinfo{volume}{32}, \bibinfo{number}{4} (\bibinfo{year}{2013}),
  \bibinfo{pages}{46:1--46:12}.
\newblock


\bibitem[\protect\citeauthoryear{Meyer, Wang, Zimmer, Grosse, and
  Sorkine-Hornung}{Meyer et~al\mbox{.}}{2015}]%
        {meyer2015phase}
\bibfield{author}{\bibinfo{person}{Simone Meyer}, \bibinfo{person}{Oliver
  Wang}, \bibinfo{person}{Henning Zimmer}, \bibinfo{person}{Max Grosse}, {and}
  \bibinfo{person}{Alexander Sorkine-Hornung}.}
  \bibinfo{year}{2015}\natexlab{}.
\newblock \showarticletitle{Phase-Based Frame Interpolation for Video}. In
  \bibinfo{booktitle}{{\em IEEE Conference on Computer Vision and Pattern
  Recognition (CVPR)}}. \bibinfo{pages}{1410--1418}.
\newblock


\bibitem[\protect\citeauthoryear{Mitra and Veeraraghavan}{Mitra and
  Veeraraghavan}{2012}]%
        {mitra2012light}
\bibfield{author}{\bibinfo{person}{Kaushik Mitra} {and} \bibinfo{person}{Ashok
  Veeraraghavan}.} \bibinfo{year}{2012}\natexlab{}.
\newblock \showarticletitle{Light field denoising, light field superresolution
  and stereo camera based refocussing using a {GMM} light field patch prior}.
  In \bibinfo{booktitle}{{\em IEEE Conference on Computer Vision and Pattern
  Recognition (CVPR) Workshop}}. \bibinfo{pages}{22--28}.
\newblock


\bibitem[\protect\citeauthoryear{RayTrix}{RayTrix}{2017}]%
        {RayTrix}
\bibfield{author}{\bibinfo{person}{RayTrix}.} \bibinfo{year}{2017}\natexlab{}.
\newblock \bibinfo{title}{3{D} Light Field Camera Technology}.
\newblock \bibinfo{howpublished}{\url{https://www.raytrix.de/}}.
  (\bibinfo{year}{2017}).
\newblock


\bibitem[\protect\citeauthoryear{{Red Camera}}{{Red Camera}}{2017}]%
        {red}
\bibfield{author}{\bibinfo{person}{{Red Camera}}.}
  \bibinfo{year}{2017}\natexlab{}.
\newblock \bibinfo{title}{Red Digital Cinema Camera}.
\newblock \bibinfo{howpublished}{\url{http://www.red.com/}}.
  (\bibinfo{year}{2017}).
\newblock


\bibitem[\protect\citeauthoryear{Revaud, Weinzaepfel, Harchaoui, and
  Schmid}{Revaud et~al\mbox{.}}{2015}]%
        {revaud2015epicflow}
\bibfield{author}{\bibinfo{person}{Jerome Revaud}, \bibinfo{person}{Philippe
  Weinzaepfel}, \bibinfo{person}{Zaid Harchaoui}, {and}
  \bibinfo{person}{Cordelia Schmid}.} \bibinfo{year}{2015}\natexlab{}.
\newblock \showarticletitle{EpicFlow: Edge-preserving interpolation of
  correspondences for optical flow}. In \bibinfo{booktitle}{{\em IEEE
  Conference on Computer Vision and Pattern Recognition (CVPR)}}.
  \bibinfo{pages}{1164--1172}.
\newblock


\bibitem[\protect\citeauthoryear{Sawhney, Guo, Hanna, Kumar, Adkins, and
  Zhou}{Sawhney et~al\mbox{.}}{2001}]%
        {sawhney2001hybrid}
\bibfield{author}{\bibinfo{person}{Harpreet~S Sawhney}, \bibinfo{person}{Yanlin
  Guo}, \bibinfo{person}{Keith Hanna}, \bibinfo{person}{Rakesh Kumar},
  \bibinfo{person}{Sean Adkins}, {and} \bibinfo{person}{Samuel Zhou}.}
  \bibinfo{year}{2001}\natexlab{}.
\newblock \showarticletitle{Hybrid stereo camera: an {IBR} approach for
  synthesis of very high resolution stereoscopic image sequences}. In
  \bibinfo{booktitle}{{\em ACM SIGGRAPH}}. \bibinfo{pages}{451--460}.
\newblock


\bibitem[\protect\citeauthoryear{Shi, Hassanieh, Davis, Katabi, and Durand}{Shi
  et~al\mbox{.}}{2014}]%
        {shi2014light}
\bibfield{author}{\bibinfo{person}{Lixin Shi}, \bibinfo{person}{Haitham
  Hassanieh}, \bibinfo{person}{Abe Davis}, \bibinfo{person}{Dina Katabi}, {and}
  \bibinfo{person}{Fredo Durand}.} \bibinfo{year}{2014}\natexlab{}.
\newblock \showarticletitle{Light Field Reconstruction Using Sparsity in the
  Continuous Fourier Domain}.
\newblock \bibinfo{journal}{{\em ACM Transactions on Graphics (TOG)\/}}
  \bibinfo{volume}{34}, \bibinfo{number}{1} (\bibinfo{year}{2014}),
  \bibinfo{pages}{12:1--12:13}.
\newblock


\bibitem[\protect\citeauthoryear{Wang, Efros, and Ramamoorthi}{Wang
  et~al\mbox{.}}{2015}]%
        {wang2015occlusion}
\bibfield{author}{\bibinfo{person}{Ting-Chun Wang}, \bibinfo{person}{Alexei~A
  Efros}, {and} \bibinfo{person}{Ravi Ramamoorthi}.}
  \bibinfo{year}{2015}\natexlab{}.
\newblock \showarticletitle{Occlusion-aware Depth Estimation Using Light-field
  Cameras}. In \bibinfo{booktitle}{{\em IEEE International Conference on
  Computer Vision (ICCV)}}. \bibinfo{pages}{3487--3495}.
\newblock


\bibitem[\protect\citeauthoryear{Wang, Srikanth, and Ramamoorthi}{Wang
  et~al\mbox{.}}{2016b}]%
        {wang2016semi}
\bibfield{author}{\bibinfo{person}{Ting-Chun Wang}, \bibinfo{person}{Manohar
  Srikanth}, {and} \bibinfo{person}{Ravi Ramamoorthi}.}
  \bibinfo{year}{2016}\natexlab{b}.
\newblock \showarticletitle{Depth from semi-calibrated stereo and defocus}. In
  \bibinfo{booktitle}{{\em IEEE Conference on Computer Vision and Pattern
  Recognition (CVPR)}}. \bibinfo{pages}{3717--3726}.
\newblock


\bibitem[\protect\citeauthoryear{Wang, Zhu, Hiroaki, Chandraker, Efros, and
  Ramamoorthi}{Wang et~al\mbox{.}}{2016c}]%
        {wang20164d}
\bibfield{author}{\bibinfo{person}{Ting-Chun Wang}, \bibinfo{person}{Jun-Yan
  Zhu}, \bibinfo{person}{Ebi Hiroaki}, \bibinfo{person}{Manmohan Chandraker},
  \bibinfo{person}{Alexei~A Efros}, {and} \bibinfo{person}{Ravi Ramamoorthi}.}
  \bibinfo{year}{2016}\natexlab{c}.
\newblock \showarticletitle{A 4D light-field dataset and CNN architectures for
  material recognition}. In \bibinfo{booktitle}{{\em European Conference on
  Computer Vision (ECCV)}}. \bibinfo{pages}{121--138}.
\newblock


\bibitem[\protect\citeauthoryear{Wang, Liu, Heidrich, and Dai}{Wang
  et~al\mbox{.}}{2016a}]%
        {wang2016light}
\bibfield{author}{\bibinfo{person}{Yuwang Wang}, \bibinfo{person}{Yebin Liu},
  \bibinfo{person}{Wolfgang Heidrich}, {and} \bibinfo{person}{Qionghai Dai}.}
  \bibinfo{year}{2016}\natexlab{a}.
\newblock \showarticletitle{The Light Field Attachment: Turning a {DSLR} into a
  Light Field Camera Using a Low Budget Camera Ring}.
\newblock \bibinfo{journal}{{\em IEEE Transactions on Visualization and
  Computer Graphics (TVCG)\/}} (\bibinfo{year}{2016}).
\newblock


\bibitem[\protect\citeauthoryear{Wanner and Goldluecke}{Wanner and
  Goldluecke}{2014}]%
        {wanner2014variational}
\bibfield{author}{\bibinfo{person}{Sven Wanner} {and} \bibinfo{person}{Bastian
  Goldluecke}.} \bibinfo{year}{2014}\natexlab{}.
\newblock \showarticletitle{Variational Light Field Analysis for Disparity
  Estimation and Super-Resolution}.
\newblock \bibinfo{journal}{{\em IEEE Transactions on Pattern Analysis and
  Machine Intelligence (TPAMI)\/}} \bibinfo{volume}{36}, \bibinfo{number}{3}
  (\bibinfo{year}{2014}), \bibinfo{pages}{606--619}.
\newblock


\bibitem[\protect\citeauthoryear{Wilburn, Smulski, Lee, and Horowitz}{Wilburn
  et~al\mbox{.}}{2002}]%
        {wilburn2002light}
\bibfield{author}{\bibinfo{person}{Bennett Wilburn}, \bibinfo{person}{Michael
  Smulski}, \bibinfo{person}{HH~Kelin Lee}, {and} \bibinfo{person}{Mark
  Horowitz}.} \bibinfo{year}{2002}\natexlab{}.
\newblock \showarticletitle{The light field video camera}. In
  \bibinfo{booktitle}{{\em SPIE Proc. Media Processors}},
  Vol.~\bibinfo{volume}{4674}.
\newblock


\bibitem[\protect\citeauthoryear{Wu, Zhao, Wang, Dai, Chai, and Liu}{Wu
  et~al\mbox{.}}{2017}]%
        {wu2017light}
\bibfield{author}{\bibinfo{person}{Gaochang Wu}, \bibinfo{person}{Mandan Zhao},
  \bibinfo{person}{Liangyong Wang}, \bibinfo{person}{Qionghai Dai},
  \bibinfo{person}{Tianyou Chai}, {and} \bibinfo{person}{Yebin Liu}.}
  \bibinfo{year}{2017}\natexlab{}.
\newblock \showarticletitle{Light Field Reconstruction Using Deep Convolutional
  Network on {EPI}}. In \bibinfo{booktitle}{{\em IEEE Conference on Computer
  Vision and Pattern Recognition (CVPR)}}.
\newblock


\bibitem[\protect\citeauthoryear{Yoon, Jeon, Yoo, Lee, and So~Kweon}{Yoon
  et~al\mbox{.}}{2015}]%
        {yoon2015learning}
\bibfield{author}{\bibinfo{person}{Youngjin Yoon}, \bibinfo{person}{Hae-Gon
  Jeon}, \bibinfo{person}{Donggeun Yoo}, \bibinfo{person}{Joon-Young Lee},
  {and} \bibinfo{person}{In So~Kweon}.} \bibinfo{year}{2015}\natexlab{}.
\newblock \showarticletitle{Learning a Deep Convolutional Network for
  Light-Field Image Super-Resolution}. In \bibinfo{booktitle}{{\em IEEE
  International Conference on Computer Vision (ICCV) Workshop}}.
  \bibinfo{pages}{57--65}.
\newblock


\bibitem[\protect\citeauthoryear{Zhang, Liu, and Dai}{Zhang
  et~al\mbox{.}}{2015}]%
        {zhang2015light}
\bibfield{author}{\bibinfo{person}{Zhoutong Zhang}, \bibinfo{person}{Yebin
  Liu}, {and} \bibinfo{person}{Qionghai Dai}.} \bibinfo{year}{2015}\natexlab{}.
\newblock \showarticletitle{Light field from micro-baseline image pair}. In
  \bibinfo{booktitle}{{\em IEEE Conference on Computer Vision and Pattern
  Recognition (CVPR)}}. \bibinfo{pages}{3800--3809}.
\newblock


\bibitem[\protect\citeauthoryear{Zhou, Tulsiani, Sun, Malik, and Efros}{Zhou
  et~al\mbox{.}}{2016}]%
        {zhou2016view}
\bibfield{author}{\bibinfo{person}{Tinghui Zhou}, \bibinfo{person}{Shubham
  Tulsiani}, \bibinfo{person}{Weilun Sun}, \bibinfo{person}{Jitendra Malik},
  {and} \bibinfo{person}{Alexei~A Efros}.} \bibinfo{year}{2016}\natexlab{}.
\newblock \showarticletitle{View Synthesis by Appearance Flow}. In
  \bibinfo{booktitle}{{\em European Conference on Computer Vision (ECCV)}}.
  \bibinfo{pages}{286--301}.
\newblock


\end{thebibliography}

\end{document}